\documentclass[10pt,journal,letterpaper,compsoc]{IEEEtran}
\usepackage{graphicx}
\usepackage{amsmath}
\usepackage{amssymb}
\usepackage{floatflt}
\usepackage{subfigure}
\usepackage{algorithm}
\usepackage{algorithmic}
\usepackage{color}

\graphicspath{{figs/}{pics1/}{./}}
\DeclareGraphicsExtensions{.jpg,.pdf,.mps,.png}

\def\argmin{\mathop{\rm argmin}}
\def\RR{\mathbb R}

\newcommand{\bx}{\boldsymbol{x}}
\newcommand{\bu}{\boldsymbol{u}}
\newcommand{\bbeta}{{\boldsymbol{\beta}}}
\newcommand{\be}{{\boldsymbol{\epsilon}}}
\newcommand{\bsb}{\boldsymbol}

\newcommand{\bsbX}{{\boldsymbol{X}}}
\newcommand{\bsbx}{{\boldsymbol{x}}}

\newcommand{\bsbV}{{\boldsymbol{V}}}

\newtheorem{theorem}{Theorem}[section]
\newtheorem{lemma}[theorem]{Lemma}

\title{Feature Selection with Annealing for Computer Vision and Big Data Learning}

\author{Adrian Barbu,  Yiyuan She, Liangjing Ding, Gary Gramajo
\IEEEcompsocitemizethanks{\IEEEcompsocthanksitem A. Barbu, Y. She and G. Gramajo are with the Department of Statistics, Florida State University, Tallahassee, Florida 32306, USA, Fax: 850-644-5271, Email: abarbu@stat.fsu.edu, yshe@stat.fsu.edu, ggramajo@stat.fsu.edu. This work was supported in part by DARPA MSEE grant FA 8650-11-1-7149, DARPA SIMPLEX N66001-15-C-4035 and NSF grant DMS-1352259.

}\vspace{-1mm}}

\begin{document}

\maketitle

\begin{abstract}
Many computer vision and medical imaging problems are faced with learning from large-scale datasets, with millions of observations and features. In this paper we propose a novel efficient learning scheme that tightens a sparsity constraint by gradually removing variables based on a criterion and a schedule. The attractive fact that the problem size keeps dropping throughout the iterations makes it particularly suitable for big data learning. Our approach applies generically to the optimization of any differentiable loss function, and finds  applications in regression, classification and ranking. The resultant algorithms   build variable screening  into  estimation and are extremely simple to implement. We provide theoretical guarantees of convergence and  selection consistency.  In addition, one dimensional piecewise linear response functions are used to account for nonlinearity  and a second order prior is imposed on these functions to avoid overfitting.  Experiments on real and synthetic data show that the proposed method compares very well with other state of the art methods in regression, classification and ranking while being computationally very efficient and scalable.     \vspace{-3mm}
\end{abstract}


\vspace{-3mm}
\section{Introduction}

Feature selection is a popular  and crucial technique to speed computation and to obtain parsimonious models that generalize well. Many computer vision and medical imaging problems require learning classifiers from large amounts of data, with millions of  features and even more observations. Such big data  pose great challenges for feature selection.
\begin{itemize}
\item \textbf{Efficiency.} Learning algorithms that are  fast \textit{and}  {scalable} are attractive in large-scale computation.
\item \textbf{Statistical guarantee.} In consideration of the   inevitable noise contamination and numerous nuisance dimensions in  big datasets, a trustworthy learning approach must  recover  genuine signals with high probability.
\item \textbf{Universality.} Rather than restricting to a specific problem, a universal  learning scheme can adapt to different types of problems, including, for instance, regression, classification,  ranking and others.
\item \textbf{Implementation ease.} Algorithms that are simple to implement can  avoid over-fitting and ad-hoc designs.
Regularization parameters should be defined with ease of tuning in mind.
In some real-world applications, it is helpful to have an algorithm with customizable cost based on computing resources.
\item \textbf{Nonlinearity.} Linear combinations of explanatory variables may not suffice in learning and modeling.   Incorporating nonlinearity  is  vital in many big data applications.
\end{itemize}\vspace{-2mm}

Recently, penalized methods have received a lot of attention in high-dimensional feature selection. They solve a class of  optimization problems with sparsity-inducing penalties such as the $L_1$, $L_0$, and SCAD \cite{donoho2006stable,zhang2010nearly,fan2001variable}. There is    a  statistical guarantee     that  junk dimensions can be removed with high probability (even in high dimensions) \cite{zhao2007model,zhang2010nearly}. But these optimization based algorithms are not scalable enough and the tuning of the penalty parameter could be time consuming on large datasets. Most of the these methods  cannot adaptively capture the   nonlinearity.

Boosting can also be used for feature selection when restricting each weak learner to be constructed from  a single variable. Boosting algorithms run in a progressive  manner: at each iteration a weak learner is added to the current model for the sake of  decreasing the value of a certain loss function \cite{schapire1990strength,friedman2000additive,li2004floatboost,li2010robust}.  
 What feature will be selected in the next boosting iteration strongly depends on the subset of selected features and their current coefficients. 
Such a design examines essentially all features at each boosting iteration, and since hundreds or even thousands of such iterations are usually required, boosting may not be fast enough in big data computation. 

\begin{figure*}[ht]
\centering
\includegraphics[width=0.3\linewidth]{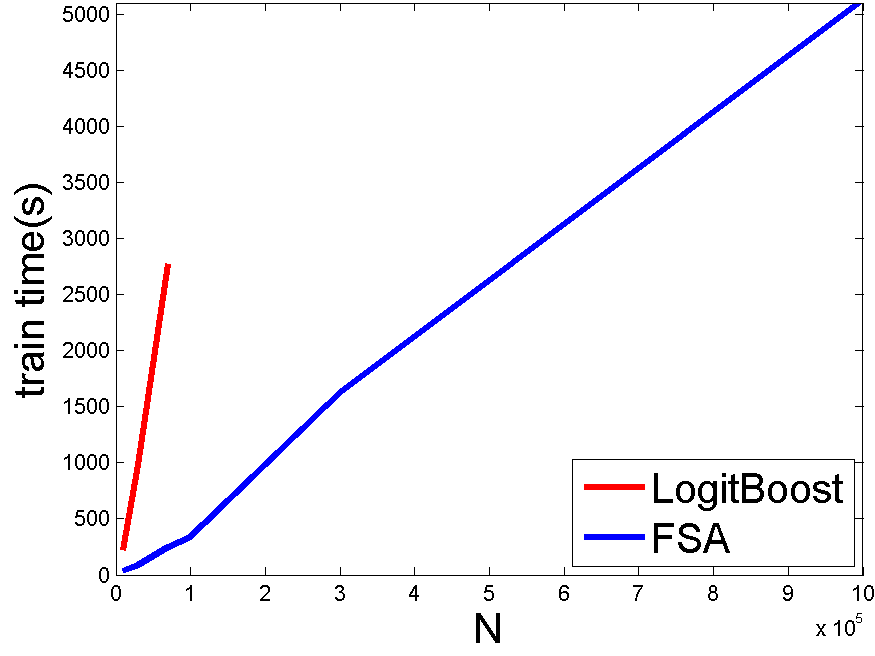}
\includegraphics[width=0.3\linewidth]{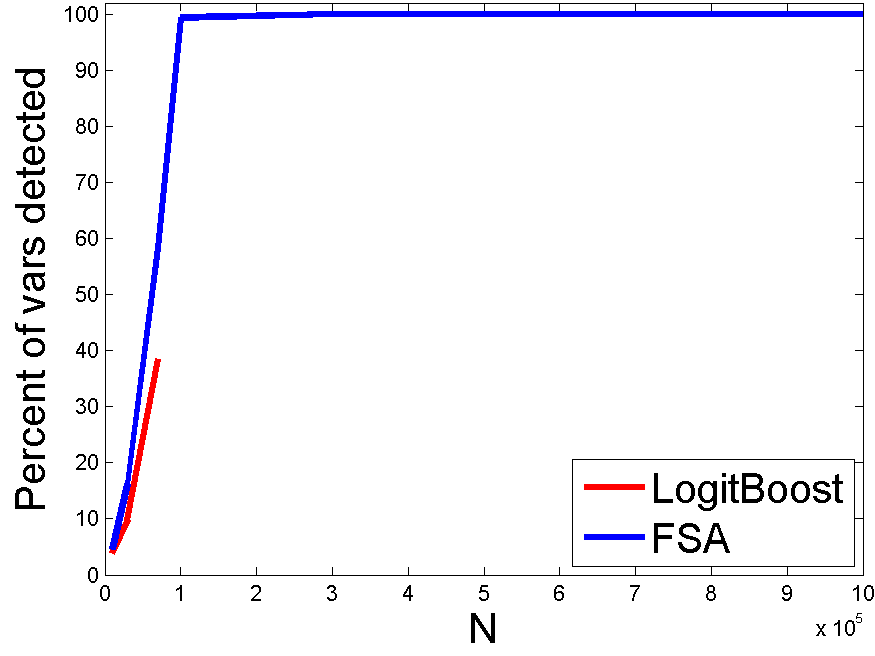}
\includegraphics[width=0.3\linewidth]{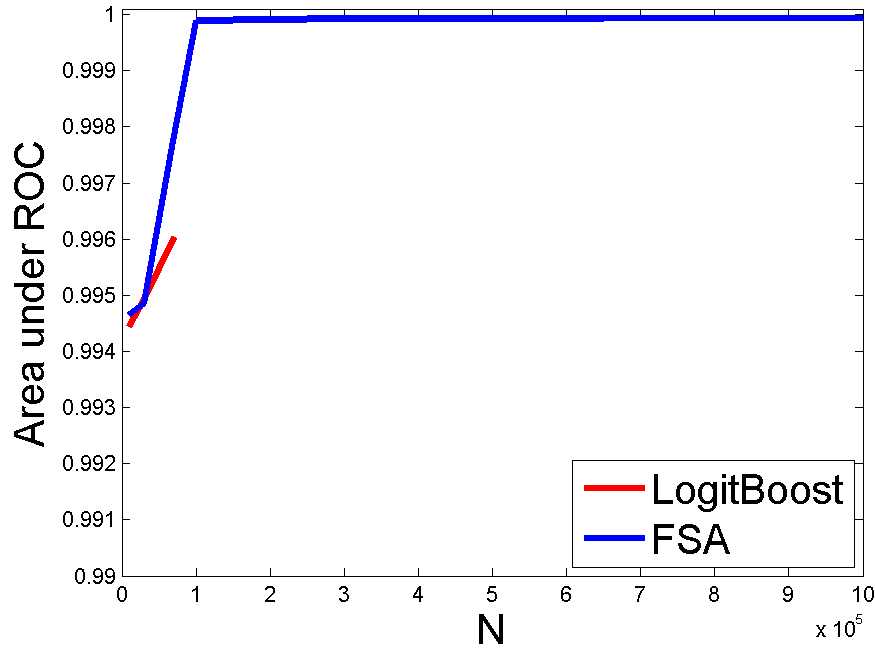}
\vskip -4mm
\caption{Classification comparison between FSA and Logitboost for equally correlated data with $M=10^5$ features and $N\leq 10^6$ observations, with $k=500$ variables selected. Left: training time. Middle: percent of variables correctly detected (see sec \ref{sec:expl1}). Right: area under ROC curve. }
\label{fig:bigdata}
\vspace{-7mm}
\end{figure*}

There also exist numerous ad-hoc procedures  designed for feature selection in specific problems. Although many ideas in this class of methods are motivating, there is a lack of   universal learning schemes that are simple to implement and can adapt to different situations.

In this paper we combine the regularization technique and the sequential algorithm design to bring forward a novel feature selection scheme that could be suitable for big data learning and has some theoretical guarantees.

Rather than growing a model by adding one variable at a time, we consider a shrinkage estimation problem in the whole predictor space, together with the use of  annealing  to lessen greediness. An attractive feature is that  a number of variables are removed while the model parameters are updated each time, which makes the problem size keep dropping during the iteration process. It is worth mentioning that our learning scheme is not ad-hoc and the principle of keep or kill has an exact form  with theoretical guarantee of optimality and consistency.

The proposed feature selection approach can handle large datasets without being online (which might be too greedy and inaccurate).
The same idea has been successfully applied previously in different areas of  signal processing \cite{She2013Spec},  reduced rank modeling \cite{She2013Mat} and network screening \cite{She2014sigmoid}.  The total amount of data the algorithm needs to access  for training is about 2 to 10 times the size of the training set, which can be orders of magnitude faster than penalization or Boosting algorithms. The algorithm can be easily  distributed over a grid of processors for even larger scale problems.

Experiments on extensive  synthetic and real data (including \textit{face keypoint detection} and  \textit{motion segmentation}) provide empirical evidence that the proposed FSA learning has performance   comparable to  or better than up-to-date penalization  and boosting methods while runs much more efficiently  on large datasets.

\vspace{-2mm}
\subsection{Some Related Works}
\label{sec:relworks}

We briefly discuss some related works to FSA, which are grouped in feature selection methods, penalized loss algorithms and boosting.

\noindent{\bf Feature Selection Methods.}
FSA shares some similarity to the Recursive Feature Elimination \cite{guyon2002gene} (RFE) procedure, which alternates training an SVM classifier on the current feature set and removing a percentage of the features based on the magnitude of the variable coefficients. However, our approach has the following significant differences:

\begin{enumerate}
\item It removes numerous  junk variables long before the parameters $\bbeta$  have converged, thus being much faster than the RFE approach where all coefficients are fully trained at each iteration.
\item It can be applied to any loss function, not necessarily the SVM loss and we present applications in regression, classification and ranking.
\item It offers rigorous theoretical guarantees of variable selection and parameter consistency.
\end{enumerate}\vspace{-1mm}

FSA can be viewed as a backward elimination method  \cite{guyon2006feature}. But its variable elimination is built into the optimization process. Although there are many methods for  variable removal and model update,  our algorithm design of combining the optimization update and progressive killing is  unique to the best of our knowledge. These principles enjoy theoretical guarantees of convergence, variable selection and parameter consistency.

There exist feature selection methods such as MRMR \cite{peng2005feature} and Parallel FS \cite{zhou2014parallel} that only select features, independent of the model that will be built on those features. In contrast, our method simultaneously selects features and builds the model on the selected features in a unified approach aimed at minimizing a loss function with sparsity constraints.

\noindent {\bf Penalized loss algorithms}  add a sparsity inducing penalty such as the $L_1$\cite{bunea2007sparsity,donoho2006stable,knight2000asymptotics,zhao2007model}, SCAD\cite{fan2001variable},  MCP\cite{zhang2010nearly} and the $L_0+L_2$ \cite{SheTISP,SheGLMTISP} and optimize  a non-differentiable objective  loss function in various ways.
The proposed method is different from the penalized methods because variable selection is not obtained by imposing a sparsity prior on the variables, but by a successive optimization and reduction of the $L_0$ constrained loss function. The sparsity parameter $k$ in FSA is more intuitive than penalty parameters and provides direct cardinality control of the obtained model.

FSA does not introduce any undesired bias on the coefficients.
 In contrast, the bias introduced by the $L_1$ penalty for a certain sparsity level might be too large and it can lead to poor classification performance \cite{bruce1996understanding,efron2004least,fan2001variable}. This is why it is a common practice when using the $L_1$ penalty to fit the penalized model only for variable selection and to refit an unpenalized model on the selected variables afterwards. Such a two-step procedure is not necessary in the approach proposed in this paper.

Another related class of methods are based on Stochastic Gradient Descent, such as \cite{langford2009sparse,tsuruoka2009stochastic,xiao2010dual,zhang2004solving}. However, they still use a sparsity inducing penalty to obtain feature selection, which makes it difficult to optimize and can be slow in practice. We will present in Section \ref{sec:expl1} an evaluation of an implementation of \cite{tsuruoka2009stochastic} and see that it lags behind our method in  computation time, feature selection accuracy and prediction power.

\noindent{\bf Boosting.} It would also be interesting to compare FSA with boosting. Boosting algorithms -- such as Adaboost \cite{schapire1990strength}, Logitboost \cite{friedman2000additive}, Floatboost \cite{li2004floatboost}, Robust Logitboost \cite{li2010robust} to cite only a few -- optimize a loss function in a greedy manner in $k$ iterations, at each iteration adding a weak learner that decreases the loss most. There are other modern versions such as LP-Adaboost \cite{grove1998boosting}, arc-gv \cite{breiman1999prediction}, Adaboost* \cite{ratsch2002maximizing}, LP-Boost \cite{demiriz2002linear},  Optimal Adaboost \cite{rudin2004dynamics}, Coordinate Ascent Boosting and and Approximate Coordinate Ascent Boosting \cite{rudin2004boosting}, which aim at optimizing a notion of the margin at each iteration. Boosting has been regarded as a coordinate descent algorithm \cite{rosset2004boosting,rudin2004boosting} optimizing a loss function that can be margin based.

Boosting algorithms do not explicitly enforce sparsity but can be used for feature selection by using weak learners that depend on a single variable (feature). What feature will be selected in the next boosting iteration depends on what features have already been selected and their current coefficients. This dependence structure makes it difficult to obtain  a general theoretical variable selection guarantees for boosting.

The approach introduced in this paper is different from boosting because it starts with all the variables and gradually removes variables, according to an elimination schedule. Indeed, its top-down design is opposite to that of boosting, but seems to be less greedy in feature selection based on our experiments in Section \ref{sec:results}. Perhaps more importantly, in  computation, FSA does not have to rank \textit{all} features at each step but keeps dropping the problem size, and the total cost can be customized based on computation resources.

\vspace{-2mm}
\section{The Feature Selection with Annealing Algorithm} \label{sec:fsa}

Let $(\bx_i,y_i),i=\overline{1,N}$ be training examples with $\bx_i\in \RR^M$ and a loss function $L(\bbeta)$ defined based on these examples.
We formulate the feature selection problem as  a \textbf{constrained} optimization
\vspace{-2mm}
\begin{equation}
\bbeta=\argmin_{|\{j:\beta_j\not =0\}|\leq k} L(\bbeta)
\label{eq:loss}
\vspace{-2mm}
\end{equation}
where the number $k$ of relevant features is a given parameter, and the loss function $L(\bbeta)$ is differentiable with respect to $\bbeta$.
This constrained form facilitates parameter tuning because  in comparison with  penalty parameters such as  $\lambda$ in $\lambda \| \boldsymbol{\beta}\|_1$, our regularization parameter $k$ is much more intuitive   and easier to specify.  The experiments in Section  \ref{sec:stability} also demonstrate  the robustness of the choice of $k$ as long as it is within a large range.
Of course, with such a   nonconvex (and discrete) constraint, the optimization problem is challenging to solve especially for large     $M$.

\vspace{-2mm}
\subsection{Basic Algorithm Description}

Our key ideas in the algorithm design  are:  a) using an annealing plan to lessen  the greediness in reducing the dimensionality from $M$ to $k$, and b) gradually removing the most irrelevant variables to facilitate computation.
The prototype algorithm summarized in Algorithm \ref{alg:fsa1} is actually pretty simple. It starts with an initial value of the parameter vector $\bbeta$, usually $\bbeta=0$, and alternates two basic steps:
one step of parameter updates towards minimizing the loss $L(\bbeta)$ by gradient descent
\vspace{-2mm}
\begin{equation}
\bbeta\leftarrow \bbeta-\eta \frac{\partial L(\bbeta)}{\partial \bbeta}, \label{eq:betaupd}
\vspace{-2mm}
\end{equation}
and one variable selection step that removes some variables according to the coefficient magnitudes  $|\beta_j|,j=\overline {1,M}$.
\vspace{-4mm}
\begin{algorithm}[htb]
   \caption{{\bf Feature Selection with Annealing (FSA)}}
   \label{alg:fsa1}
\begin{algorithmic}
   \STATE {\bfseries Input:} Training examples $\{(\bx_i,y_i)\}_{i=1}^N$.
   \STATE {\bfseries Output:} Trained classifier parameter vector $\bbeta$.
\end{algorithmic}
\begin{algorithmic} [1]
\STATE Initialize $\bbeta=0$.
        \FOR {e=1 to $N^{iter}$}
                \STATE  Update $\bbeta \leftarrow \bbeta-\eta \frac{\partial L(\bbeta)}{\partial \bbeta}$
                \STATE Keep only the $M_e$ variables with highest $|\beta_j|$ and renumber them $1,...,M_e$.
      \ENDFOR
\end{algorithmic}
\end{algorithm}
\vspace{-4mm}

Through the annealing schedule,  the support set of the coefficient vector is gradually tightened till we reach  $|\{j,\beta_j\not =0\}|\leq k$.
Step 4 conducts an adaptive screening, resulting a nonlinear operator that increases the difficulty of the theoretical analysis. Perhaps surprisingly, the  keep-or-kill rule is simply based on the  magnitude of coefficients and  does \textit{not} involve any information of the objective function $L$.  This is in contrast to many  ad-hoc  backward elimination approaches. Nicely, Theorem \ref{thm:consist} shows such a design always has a rigorous guarantee of  computational convergence and statistical consistency.

The prototype FSA algorithm is extremely simple to implement. More importantly,  the problem size and thus the complexity keep dropping, owing to the removal process. With the annealing schedule, the nuisance features that are difficult to identify are handled only when we are close to   an optimal solution, while those `apparent' junk dimensions  are  eliminated at earlier stages to save the computational cost.
\begin{figure}[ht]
\vspace{-4mm}
\centering
\includegraphics[width=6.cm,height=3.5cm]{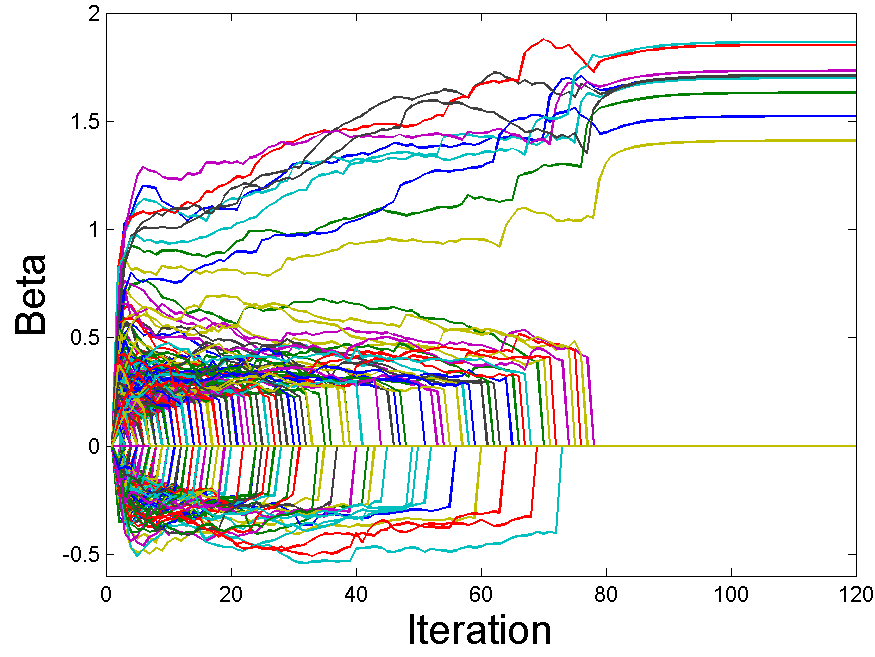}
\vskip -3mm
\caption{\small The value of $\beta_j, j=\overline{1,M}$ vs  iteration number for simulated data with $N=1,000,M=1,000,k=10$ with $\eta=20,\mu=300$. }
\vspace{-4mm}
\label{fig:beta}
\end{figure}

 Figure \ref{fig:beta} gives a demonstration of the removal and convergence process for a  classification problem with $N=1,000$ observations and  $M=1,000$ variables described in Section \ref{sec:expl1}. Notice that  some of the $\beta_j$ are zeroed after each iteration.
The algorithm stabilizes very quickly (about 80 steps for $M=1,000$).

 \begin{figure}[htb]
\vspace{-4mm}
\centering
\includegraphics[width=7.cm]{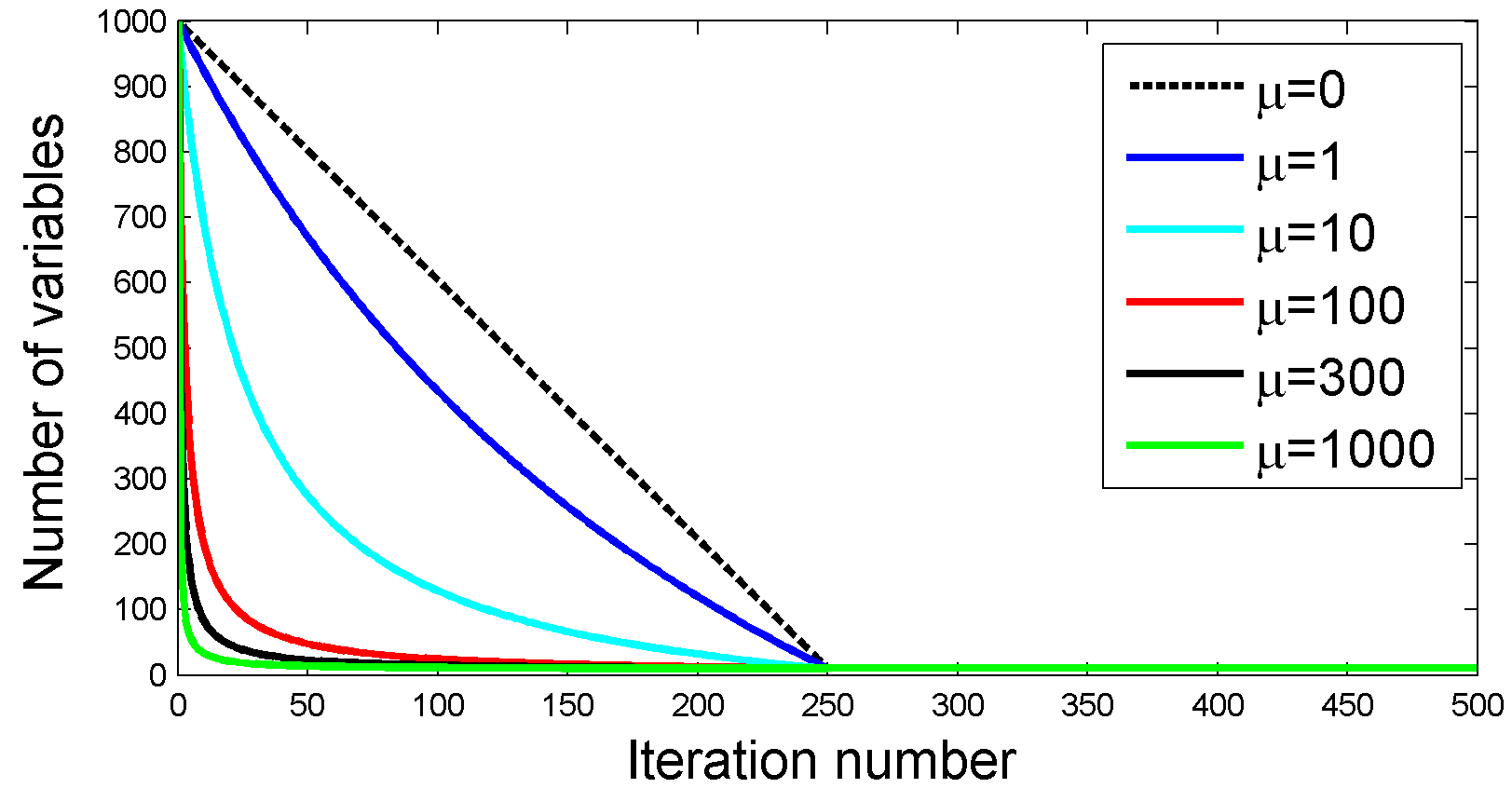}
\vskip -3mm
\caption{\small The number of kept features $M_e$ vs iteration $e$ for different  schedules, with $M=1,000, k=10, N^{iter}=500$.}
\vspace{-6mm}
\label{fig:schedule}
\end{figure}


\vspace{-2mm}
\subsection{Some Implementation Details}

In this part, we provide  empirical values of the algorithmic parameters in implementation.

First,   any annealing schedule $\{M_e\}$ slow enough  works well in terms of estimation and selection accuracy. But a fast decaying schedule could reduce the computational cost significantly. Our experience shows that the following  inverse schedule with a parameter $\mu$ provides a good balance between efficiency and accuracy:
\vspace{-2mm}
\begin{equation}
 M_e=k+(M-k)\max(0,\frac{N^{iter}- 2 e}{2  e \mu +N^{iter}}), e=\overline{1,N^{iter}}
\label{invschedule}
\vspace{-1mm}
\end{equation}

Figure \ref{fig:schedule} plots the schedules for  six difference choices of  $\mu$ with $M=1,000, k=10$ and  $N^{iter}=500$.

 The computation time is proportional to the area under the graph of the schedule curve and can be easily calculated.
Examples of computation times  are in Table \ref{tab:times}. In reality,
the overall computational complexity of FSA is \textbf{linear} in  $M N$ (the problem size).
\begin{table}[htb]
\vspace{-3mm}
\small
\begin{center}
\caption{\small{Computation times for selecting $k$ variables using $N$ observations of dimension $M$, when $N^{iter}=500$.}}\label{tab:times}
\vskip -2mm
\begin{tabular}{lc}
\hline
Annealing param $\mu$\phantom{$I^I$} &Computation Time\\
\hline
$\mu=0$\phantom{$I^{I^I}$} & $125MN+kNN^{iter}$\\
$\mu=1$ & $97MN+kNN^{iter}$\\
$\mu=10$ & $41MN+kNN^{iter}$\\
$\mu=100$ & $10MN+kNN^{iter}$\\
$\mu=300$ & $5MN+kNN^{iter}$\\
$\mu=1000$ & $2MN+kNN^{iter}$\\
\hline
\end{tabular}
\end{center}
\vspace{-6mm}
\end{table}

In addition to the annealing schedule, the performance of FSA depends on two other parameters:
\begin{itemize}
\item Gradient learning rate $\eta$, which can be arbitrarily small provided that the number of iterations is large enough. Of course, if $\eta$ is too large, the coefficients $\beta_j$ may not converge. We used $\eta=20$ for classification and $\eta=1$ for regression.
\item Number of iterations $N^{iter}$, large enough to insure the parameters have converged to a desired tolerance. In our experiments we used $N^{iter}=500$.
\end{itemize}


Finally, we observe that the performance of the algorithm is rather stable for a large range of values for the parameters $\eta,\mu, N^{iter}$ (cf.  Section \ref{sec:stability}). This is advantageous in implementation and parameter tuning.

\noindent {\bf Large Scale Implementation.} The FSA algorithm can be parallelized for large scale problems by subdividing the $N\times M$ data matrix into a grid of sub-blocks that fit into the memory of the processing units. Then the per-observation response vectors can be obtained from a row-wise reduction of the partial sums computed by the units. The parameter updates are done similarly, via column-wise reduction. A GPU  based implementation could offer further computation cost reductions.

\vspace{-2mm}
\subsection{Examples and Variants}\label{sec:losses}
The FSA algorithm can be used for the optimization of any differentiable loss function subject with a sparsity constraint as described in eq. \eqref{eq:loss}. Some examples are given as follows in  regression, classification and  ranking.

\noindent{\textbf{FSA for Regression.}}
Given training examples $(\bx_i,y_i)\in \RR^M\times \RR,i=\overline{1,N}$, we have the penalized squared-error loss
\vspace{-2mm}\begin{equation}
L(\bbeta)=\sum_{i=1}^N  (y_i-\bx_{i}^T\bbeta)^2
+ \sum_{j=1}^M\rho(\beta_j)
\vspace{-2mm} \label{eq:penloss}
\end{equation}
with a differentiable prior function $\rho$ such as $\rho(\beta)=s\beta^2$.

\begin{figure}[ht]
\vspace{-3mm}
\centering
\includegraphics[width=4.3cm]{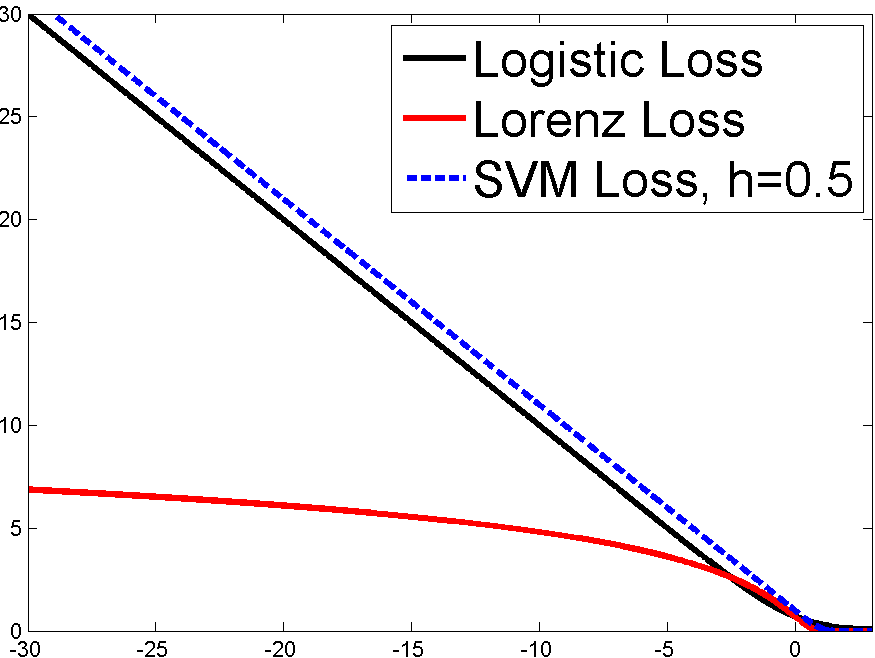}
\includegraphics[width=4.3cm]{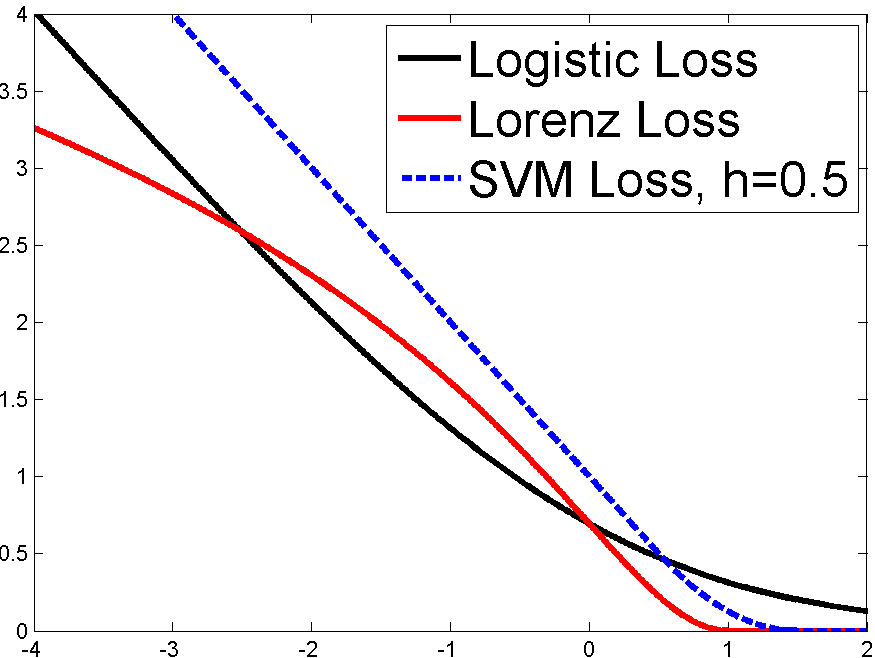}
\vskip -2mm
\caption{The loss functions from eq. \eqref{eq:logloss}, \eqref{eq:svmloss} and \eqref{eq:lorloss}. Left: the losses on the interval $[-30,3]$. Right: zoom in the interval $[-4,2]$.}
\vspace{-3mm}
\label{fig:losses}
\end{figure}
\noindent{\textbf{FSA for Classification.}}
FSA can be used for classification and  feature selection.
Given a set of training examples $D=\{(\bx_i,y_i)\in \RR^M\times \{-1,1\},i=\overline{1,N}\}$ the loss functions
\vspace{-3mm}\begin{equation}
L_D(\bbeta)=\sum_{i=1}^N w_i \ell(y_i \bx_i^T\bbeta) + \sum_{j=1}^M\rho(\bbeta_j)
\vspace{-2.5mm} \label{eq:clfloss}
\end{equation}
are based on the per-example loss functions $\ell:\RR \to \RR$ illustrated in Figure \ref{fig:losses}:
\begin{itemize}
\item {The Logistic Loss} is
\vspace{-3mm}\begin{equation}
\ell(x)=\ln(1+\exp(- x))
\vspace{-2mm} \label{eq:logloss}
\end{equation}

\item {The SVM Loss} has $w_i=1,i=\overline{1,N}$  and
is based on the Huber-style differentiable approximation of the hinge loss \cite{chapelle2007training}:
\vspace{-3mm}
\begin{equation}
\ell_h(x)=\begin{cases}
0 &\text{ if } x>1+h\\
\frac{\displaystyle{(1+h-x)^2}}{\displaystyle{4h}} &\text{ if } |1-x|\leq h\\
1-x &\text{ if }x<1-h
\end{cases} \label{eq:svmloss}
\vspace{-2mm}
\end{equation}

\item {The Lorenz Loss} is a novel loss function we introduce in this paper, also using $w_i=1,i=\overline{1,N}$
\vspace{-2mm}
\begin{equation}
\ell(x)=\begin{cases}
0 &\text{ if } x>1\\
\ln(1+(x-1)^2) &\text{ else}
\end{cases} \label{eq:lorloss}
\vspace{-2mm}
\end{equation}
 \end{itemize}
The Lorenz loss is differentiable everywhere, it is zero for $x\in [1,\infty)$ and grows logarithmically with respect to $|x|$  as $x\to -\infty$. These properties make the Lorenz loss \eqref{eq:lorloss} behave like the SVM loss in the sense that correctly classified examples that are far from the margin don't contribute to the loss. Moreover, the Lorenz loss is more robust to label noise than the SVM and logistic losses because the loss values for the misclassified examples that are far from the margin is not much higher than for those that are close to the margin. This loss is not convex, but it works well in practice together with the FSA algorithm, as it will be seen in experiments.

\noindent{\textbf{FSA for Ranking.}}
We developed an  extension of FSA to deal with ranking problems. Let $\bx_i\in \RR^M$ be the training instances and $r_{ij}\in [0,1]$ be the true rankings between observations $\bx_i,\bx_j$, for some pairs $(i,j)\in C\subset \{1,...,N\}\times \{1,...,N\}$. A criterion (e.g. an error measure) can be used to compare instances $\bx_i$ and $\bx_j$ and generate the true rankings $r_{ij}\in [0,1]$, which can be for example $0$ if $\bx_i$ is "better" than $\bx_j$, $0.5$ if they are "equally good" and $1$ if $\bx_i$ is "worse" than $\bx_j$.

Here, training refers to finding a ranking function $f_\bbeta(\bx): \RR^M\to \RR$ specified by a parameter vector $\bbeta$ such that $f_\bbeta(\bx_i)-f_\bbeta(\bx_j)$ agrees as much as possible with the true rankings $r_{ij}$.

There are different criteria that could be optimized to measure this degree of agreement. Motivated by  \cite{burges2005learning} we adopt a  differentiable criterion with a prior term $\sum_{j=1}^M\rho(\beta_j)$
\vspace{-2.5mm}\begin{equation}
\vspace{-1.5mm}\begin{split}
L_C(\bbeta)&=\sum_{(i,j)\in C}\ln [1+\exp(\bx_i^T\bbeta-\bx_j^T\bbeta)]\\  \label{eq:rankloss}
&-\sum_{(i,j)\in C} r_{ij}(\bx_i^T\bbeta-\bx_j^T\bbeta)+ \sum_{j=1}^M\rho(\beta_j)
\end{split}
\end{equation}
where  $f_\bbeta(\bx)=\bx^T\bbeta$.

\vspace{-2mm}
\subsection{Convergence and Consistency Theorem}
\label{sec:th}
We investigate the performance of the FSA estimators  in  {regression} and {classification} problems. In the first case, the statistical assumption is that each $y_i$ follows a Gaussian distribution $\mathcal N(\bx_i^T \bbeta^*, \sigma^2 I)$, while in the latter situation each $y_i$ is 0 or 1 following the Bernoulli distribution with mean $\exp(\bx_i^T \bbeta^*)/(1+\exp(\bx_i^T \bbeta^*))$, where  $\bbeta^*$ denotes the true coefficient vector (unknown). We focus on the log-likelihood based  loss  (denoted by $F$) in this subsection, which is  the squared-error loss  and  the logistic loss from \eqref{eq:penloss} and   \eqref{eq:logloss}, respectively. For clarity, we redefine them as follows:
\vspace{-2mm}
\begin{align}
\mbox{{Regression}: } & F(\bbeta)=\frac{1}{2}\sum_{i=1}^N  (y_i-\bx_{i}^T \bbeta)^2, \label{regloss}\\
\vspace{-4mm}
\mbox{{Classification}: } &F(\bbeta) = \hspace{-1mm}\sum_{i=1}^N (-y_i \bsbx_i^T \bbeta \hspace{-1mm}+ \hspace{-1mm} \ln(1 \hspace{-1mm}+ \hspace{-1mm}\exp(\bsbx_i^T \bbeta))).\label{clsloss}
\vspace{-6mm}
\end{align}
The FSA applications may have  $M$  large (possibly much greater than $N$).
In the rest of the subsection, we set  $N^{iter}=+\infty$ in the FSA algorithm. Let $\bbeta^{(e)}$ be the value of $\bbeta$ at iteration $e$. Let $\{M_e\}$ be a non-increasing annealing schedule satisfying $M\geq M_e \geq k$, $\forall e$ and $M_e=k$ for sufficiently large values of $e$.  Suppose   $L=F$ (for now) in either regression or classification. Define the design matrix $\bsbX=[\bsbx_1^T, \cdots, \bsbx_N^T]^T\in \mathbb R^{N\times M}$ and let  $\| \bbeta\|_0=|\{j: \beta_j\neq 0\}|$.
\begin{theorem}
\label{thm:consist}
The following convergence and consistency results hold
 under  $0<\eta < 4/\| \bsbX \|_2^2$ for  classification and  $0<\eta < 1/\| \bsbX \|_2^2$ for regression, respectively,  where $\|\bsbX\|_2$ stands for the  spectral norm of the design:

(i)  The algorithm converges  in the sense that $F(\bbeta^{(e)})$ for sufficiently large values of $e$ decreases  monotonically to a limit.

(ii) In regression (cf. \eqref{regloss}),   $\lim_{e\rightarrow \infty} \bbeta^{(e)}$ always exists; in classification (cf. \eqref{clsloss}), under the overlap condition in the appendix, the same conclusion holds. Moreover, the limit point is a locally optimal solution to $\min_{\bbeta: \|\bbeta\|_0\leq k} F(\bbeta)$.

(iii) Suppose, asymptotically,  $N\rightarrow +\infty$ and  the limit of the scaled Fisher information matrix exists, i.e., the design $\bsbX$ (or $\bsbX(N)$ as a matter of fact) and the  true coefficient $\bbeta^*$ satisfy $\lim \bsbX^T \bsbX/N \rightarrow \mathcal I^*$ in regression, or
$\lim \bsbX^T   \mbox{diag}\left\{ \frac{e^{\bsbx_i^T \bbeta^*}}{(1+e^{\bsbx_i^T \bbeta^*})^2}   \right\} \bsbX/N \rightarrow \mathcal I^*$ in classification, for some $\mathcal I^*$  positive definite. Let $ k$ be any number $ \geq \|\bbeta^*\|_0$. Then, there exists a slow enough  schedule $\{M_e\}$ such that  \emph{any} $\bbeta^{(e)}$ for $e$ sufficiently large is a consistent estimator of $\bbeta^*$, and $\{j: \beta_j^*\neq 0\}\subset \{j: \beta_j^{(e)}\neq 0\}$ occurs with probability tending to 1.
\end{theorem}

\ The proof details are given in  the supplementary material. The theorem holds  more generally for  smoothly penalized loss criteria.  For example, when    $L=F+\frac{1}{2} \lambda \|\bbeta\|_2^2$,  (i) and (ii) are true for  any $\lambda>0$, with no need of the overlap assumption in  classification.  

The convergence results, regardless of how large $k$ or $M$ can be (or even $k> N$), are reassuring  in computation.
They also imply that in implementation, we may adopt a universal choice of the stepsize at any
iteration, as long as it is properly small.
Moreover, in view of (iii), there is no need to evaluate a global  minimum (or even a local minimum).  To attain good accuracy, the cooling schedule has to be slow enough. Although coming up with  an  adaptive schedule that is theoretically sound is tempting, our current theoretical results seem to give   way too slow schedules. Based on our empirical experience we recommend using an inverse function  \eqref{invschedule} to attain a good balance of accuracy and efficiency. The optimal cooling schedule is left to further theoretical/empirical investigations in the future.

\vspace{-2mm}
\subsection{Capturing Nonlinearity in Regression, Classification and Ranking}

In this section we present methods based on the FSA technique to capture nonlinearity and structural information in the feature space in conjunction with feature selection.

We will use a type of nonlinearity that is compatible with feature selection, obtained by replacing $\bx^T\bbeta$ with a nonlinear response function that is a sum of a number of univariate functions
\vspace{-2mm}
\begin{equation}
f_\bbeta(\bx)=\displaystyle{\sum_{j=1}^M} f_{\bbeta_j}(x_j),\label{eq:nlfun}
\vspace{-2mm}
\end{equation}
where $\bbeta_j$ is a parameter vector characterizing the response function on variable $j$.

The univariate functions we will use are piecewise linear, as described in the next section.

\vspace{-3mm}
\subsubsection{Piecewise Linear Learners }\label{sec:plf}

A piecewise linear (PL) learner $f_\bbeta(x):\RR \to \RR$ is a piecewise function that only depends on one variable $x$ of the instance $\bx\in \Omega$. It is defined based on the range $[x^{min},x^{max}]$ of that variable and a predefined number $B$ of bins.

Let $b=(x^{max}-x^{min})/B$ be the bin length.
For each value $x$, the learner finds the bin index $j(x)=\left [(x-x^{min})/b \right ]\in \{0,...,B-1\}$ and the relative position in the bin $\alpha(x)=(x-x^{min})/b-j(x)\in [0,1)$ and returns
\vspace{-3mm}
\[
f_\bbeta(x)=\beta_{j(x)}(1-\alpha(x))+\beta_{j(x)+1}\alpha(x)
\vspace{-2mm}
\]
Let
\vspace{-1mm}
\[
u_{k}(x)=\begin{cases} 1-\alpha(x) & \text{ if $k=j(x)$}\\
\alpha(x) & \text{ if $k=j(x)+1$}\\
0 &\text{ else}
\end{cases}
\vspace{-2mm}
\]
for $k\in \{0,...,B\}$ be a set of $B+1$ piecewise linear basis functions. Then $f_\bbeta(x)$ can be written as a linear combination:
\vspace{-3mm}
\[
f_\bbeta(x)=\sum_{k=0}^B \beta_{k} u_{k}(x)= \bu^T(x)\bbeta
\vspace{-2.5mm}
\]
where $\bu(x)=(u_{0}(x),...,u_{B}(x))^T$ is the vector of responses of the basis functions and $\bbeta=(\beta_{0},...,\beta_{B})^T\in \RR^{B+1}$ is the parameter vector.

Some recent works \cite{Huang10,Meier09} use nonlinear additive models that depend on the variables through one dimensional smooth functions. In  \cite{Meier09} it was proved that cubic B-splines optimize a smoothness criterion on these 1D functions. Variable selection was obtained by a group lasso penalty. A similar model is presented in \cite{Ravikumar09} where a coordinate descent soft thresholding algorithm is used for optimizing an $L_1$ group-penalized loss function. Our work differs from these works by imposing constraints on the coefficients instead of biasing them with the $L_1$ penalty. Moreover, our optimization is achieved by a novel gradual variable selection algorithm that works well in practice and is computationally efficient.

\begin{figure}[ht]
\vspace{-3mm}
\centering
\includegraphics[width=5.5cm]{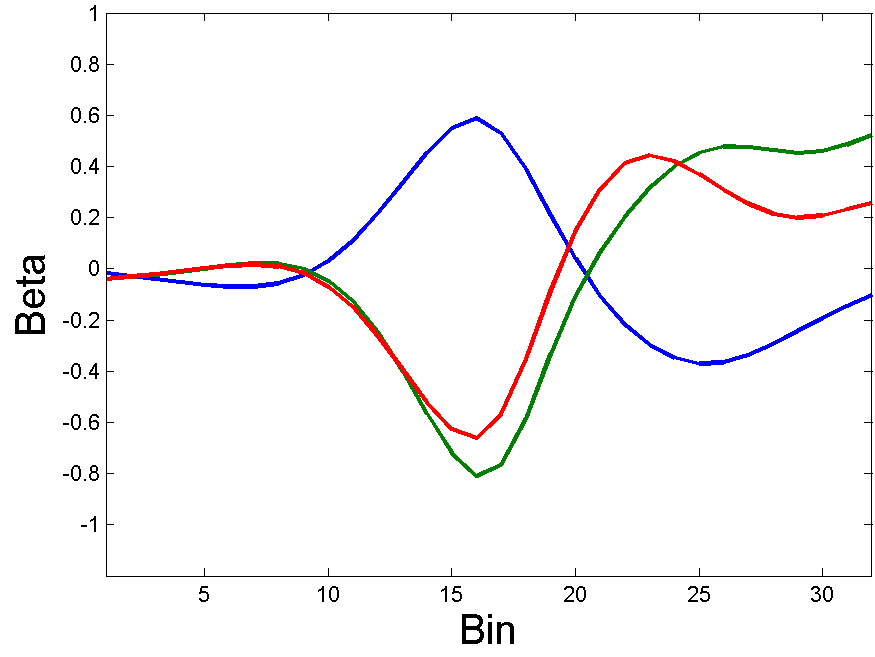}
\vskip -3mm
\caption{Piecewise linear response functions $f_{\bbeta_j}(x_j)=\bu_j^T(x_j)\bbeta_j$ obtained on an eye detection problem using the second order prior \eqref{eq:prior2}. }
\vspace{-6mm}
\label{fig:beta_2}
\end{figure}

{\bf Nonlinear Response Regularization.}
Aside from the shrinkage penalty $\rho(\bbeta_j)=\lambda\|\bbeta_j\|^2$, we will experiment with the second order prior
\vspace{-2.5mm}
\begin{equation}
\rho(\bbeta_j)=\lambda\|\bbeta_j\|^2+c \sum_{k=1}^{B-1} (\beta_{j,k+1}+\beta_{j,k-1}-2\beta_{jk})^2\label{eq:prior2}
\vspace{-2.5mm}
\end{equation}
that favors "smooth" feature response functions $h_j(\bx_j)$, as shown in Figure \ref{fig:beta_2}.

Other priors could be used, such as differentiable versions of the total variation regularization
\vspace{-3mm}
\begin{equation}
\rho(\bbeta_j)=q \sum_{k=1}^{B} h(\beta_{j,k}-\beta_{j,k-1})\label{eq:prior1}
\vspace{-2.5mm}
\end{equation}
where $h:\RR\to \RR$ could be for example the Huber approximation of the $L_1$ norm.


\vspace{-2mm}
\subsubsection{Example: Nonlinear FSA For Ranking}
\vspace{-0mm}

Using the notations from Section \ref{sec:plf}, we can use the nonlinear ranking function without intercept
\vspace{-3mm}
\begin{equation}
f_\bbeta(\bx)=\displaystyle{\sum_{j=1}^M} \bu_j^T(j_k) \bbeta_j, \label{eq:rankfn}
\vspace{-3mm}
\end{equation}
where $\bu_j(x_j)$ is the basis response vector and $\bbeta_j\in \RR^{B+1}$ is the coefficient vector of variable $j$.

For ranking we use the shrinkage prior for each coefficient vector $\bbeta_j\in \RR^{B+1}$
\vspace{-3mm}
\begin{equation}
\rho(\bbeta_j)=\lambda\|\bbeta_j\|^2, \label{eq:prior0}
\vspace{-2mm}
\end{equation}
which discourages large values of the coefficients.

The FSA-Rank method will be used in experiments to compare motion segmentations and choose the best one from a set of segmentations obtained using different parameters.

\vspace{-2mm}
\section{Experiments}
\label{sec:results}
\begin{figure*}[ht]
\centering
\includegraphics[width=0.3\linewidth]{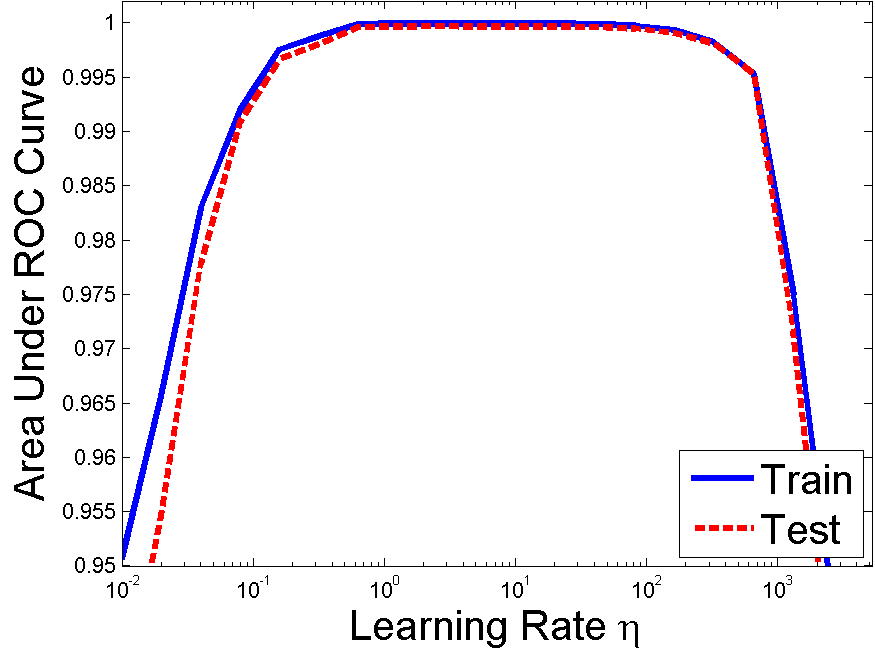}
\includegraphics[width=0.3\linewidth]{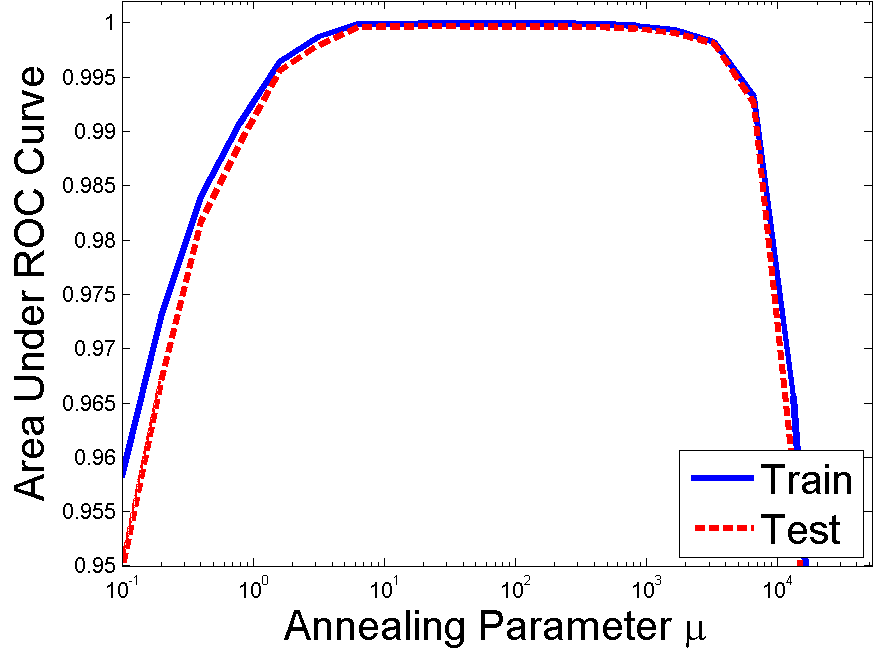}
\includegraphics[width=0.3\linewidth]{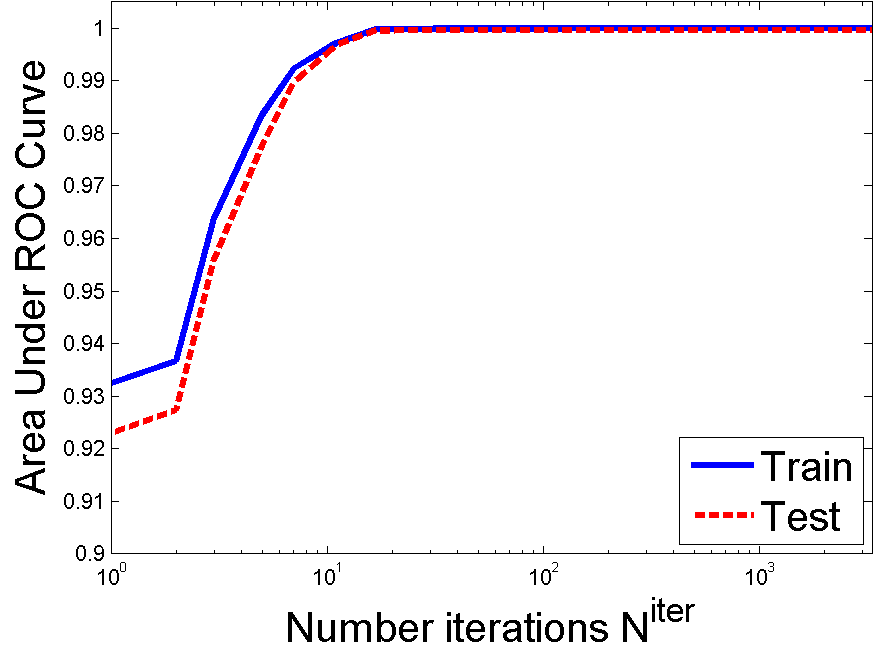}
\vskip -3mm
\caption{Dependence of the AUC vs algorithm parameters for a linear dataset with $M=N=1000, k=k^*=10$. Left: dependence on $\eta$. Middle: dependence on $\mu$ when $\eta=\mu/10$. Right: dependence on $N^{iter}$ when $\eta=20,\mu=300$.}
\label{fig:roc_eta}
\vspace{-6mm}
\end{figure*}
We first present simulations on synthetic data to evaluate feature selection and prediction performance and compare it with other state of the art feature selection methods. Then we present experiments on UCI datasets, and  applications of the classification FSA to face keypoint detection and applications of the ranking FSA to motion segmentation.

\vspace{-2mm}
\subsection{Synthetic Data Experiments}\label{sec:syndata}

In this section we focus on FSA in classification and regressions problems.
The data for simulations has correlated predictors sampled from a multivariate normal $\bx \sim {\cal N}(0,\Sigma )$ where $\Sigma_{ij}=\delta^{|i-j|}$ and $\delta=0.9$. For computational reasons, the large data experiment shown in Figure \ref{fig:bigdata} has the predictors correlated in another way, as described in Section \ref{sec:bigdata}.

For classification, the label $y$ for a data point $\bx\in \RR^M$ is
\vspace{-3mm}
\begin{equation}
y=\begin{cases} 1 &\text{ if  \;$\sum_{j=1}^{k^*}x_{10j}>0$}\\
0 &\text{ otherwise}
\end{cases}
\vspace{-2.5mm}
\end{equation}
Thus only the variables with index $10j, j=\overline{1,k}$ are relevant. We will also use a version of the data with noisy labels, where 10\% of the examples had random labels, thus about 5\% of the examples have incorrect labels.

All experiments were performed on a six core Intel Core I7-980 machine at 3.3GHz with 24Gb RAM.

\vspace{-2mm}
\subsubsection{Stability of All Algorithmic Parameters } \label{sec:stability}
\vspace{-0mm}

In this experiment, we evaluate the stability of the FSA Algorithm \ref{alg:fsa1} with respect to its tuning parameters: the learning rate $\eta$, the annealing rate $\mu$ and the number of iterations $N^{iter}$. The experiment was conducted on the linearly separable data with  $M\hspace{-1mm}=\hspace{-1mm}N\hspace{-1mm}=\hspace{-1mm}1000,$  $k\hspace{-1mm}=\hspace{-1mm}k^*\hspace{-1mm}=\hspace{-1mm}10$.

In Figure \ref{fig:roc_eta} are shown the dependence of the average area under the ROC curve (AUC) with respect to $\eta$ (left), $\mu$ (middle) and $N^{iter}$ (right). For the left plot, we had $\mu=300,N^{iter}=500$, for the middle plot $\eta=\mu/10, N^{iter}=500$ and for the right plot $\mu=300, \eta=20$. The obtained curves are the averages of 10 runs.

One can see that all three parameters have a large range of values that yield optimal prediction performance. This robustness property is in contrast to the sensitivity issue of  penalty parameters in $L_1$ or $L_0$ like methods. It greatly facilitates  parameter tuning and reduces  ad-hocness.

\vspace{-2mm}
\subsubsection{Classification Experiments} \label{sec:expl1}
\vspace{-0mm}

In this experiment, we compare the variable selection and the prediction performance of the FSA algorithm with the Logitboost algorithm and  various  sparsity-inducing penalties that are popular  in the literature. In calling Logitboost for feature selection, we require that each weak learner depend on only one variable.

\begin{table*}[htb]
\small
\begin{center}
\caption{Classification experiments on simulated linearly separable data with $\delta=0.9$, averaged over 100 runs. 
}\label{tab:l1}
\vskip -2mm
\begin{tabular}{|l|c|c|c|c|c|c|c|c|c|c|c|c|c|c|c|c|c|c|c|c|c|c|c|c|}
\hline
 \multicolumn{3}{|l|}{} &\multicolumn{11}{|c|}{\hspace{-1mm}\textbf{All-variable detection rate} (\textbf{DR})\hspace{-2mm}} &\multicolumn{11}{|c|}{\phantom{$^I$}\textbf{Percent correctly detected} (\textbf{PCD})}\\
\hline
$N$\phantom{$I^I$}\hspace{-3mm} &$M$ &$\hspace{-.5mm}k\hspace{-1mm}=\hspace{-1mm}k^*\hspace{-1mm}$    &FSA\hspace{-2mm}&FSV\hspace{0mm} &FSL  &\hspace{-0mm}QTP\hspace{-0mm}  &L1 &EL &L2 &\hspace{-.5mm}MCP\hspace{-.5mm} &\hspace{-0mm}SCD\hspace{-0mm} &\hspace{-0mm}LB\hspace{-0mm} &\hspace{-0.1mm}LB1\hspace{-1mm}  &FSA &FSV\hspace{-2mm}&FSL &QTP   &L1 &EL &L2 &MCP &SCD&LB &LB1\\
\hline
300 \phantom{$^I$} &1000 &10   & 29 &30 &34 &0   &0 &0 &0 &3 &1  &0 &0      &86.1 &84.7 &86.0  &37.9 &42.4 &40.4 &- &64.0 &55.6 &61.3 &23.1\\
1000\hspace{-2mm} &1000 &10  &100 &100 &100 &1   &2 &0 &0 &39 &25 &44  &0  &100 &100   &100  &67.8 &72.4 &49.0 &- &88.5 &85.7  &92.3 &26.3 \\
3000\hspace{-2mm} &1000 &10  &100&100 &100 &30  &33 &0 &0 &65 &63  &97  &0   &100 &100 &100 &91.1   &91.5 &60.4 &- &95.9 &95.4  &99.6 &29.1\\
10000\hspace{-2mm} &1000 &10 &100&100 &100 &88  &100 &3 &0 &97 &97  &100  &0    &100 &100 &100 &98.8   &100 &68.4 &- &99.6 &99.6  &100 &31.8\\\hline
1000 \phantom{$^I$}\hspace{-2mm} &1000 &30  &24 &22 &21 &0  &0 &0 &0 &0 &0  &0  &0          &93.8 &92.4 &92.6   &47.4 &41.5 &36.2 &- &66.8 &61.2 &62.4  &29.0\\
3000\hspace{-2mm} &1000 &30  &100 &100 &100 &0  &0 &0 &0 &8 &14 &4  &0       &100 &100 &100 &78.7 &68.6 &43.0 &- &91.1 &91.7 &90.4 &37.8\\
10000\hspace{-2mm} &1000 &30 &100 &100 &100 &33  &8 &0 &0 &73 &56 &82  &0       &100 &100 &100 &97.2 &93.9 &51.8 &- &98.3 &97.3 &99.3 &43.8\\
\hline
 \multicolumn{3}{|l|}{} &\multicolumn{11}{|c|}{\hspace{-1mm}\textbf{Area under the ROC curve} (\textbf{AUC})\hspace{-2mm}} &\multicolumn{11}{|c|}{\phantom{$^I$}\textbf{Training Time} (sec)}\\
\hline
$N$\phantom{$I^I$}\hspace{-3mm} &$M$ &$\hspace{-.5mm}k\hspace{-1mm}=\hspace{-1mm}k^*\hspace{-1mm}$    &FSA\hspace{-2mm}&FSV\hspace{-1mm} &FSL  &\hspace{-0mm}QTP\hspace{-0mm}  &L1 & EL &L2 &\hspace{-.5mm}MCP\hspace{-.5mm} &\hspace{-0mm}SCD\hspace{-0mm} &\hspace{-0mm}LB\hspace{-0mm} &\hspace{-0.1mm}LB1\hspace{-1mm}  &FSA &FSV\hspace{-2mm}&FSL &QTP   &L1 &EL &L2 &MCP &SCD&LB &LB1\\
\hline
300\phantom{$^I$}\hspace{-2mm} &1000 &10 &.992 &.990 &.990 &.899 &.915 &.937 &.922 &.955  &.934 &.950 &.923 &0.03 &0.03 &0.04 &0.02   &17 &35 &.41 &87 &68 &0.13 &0.01\\
1000 &1000 &10 &1.00  &1.00 &1.00 &.947 &.951 &.950 &.962 &.965  &.953  &.967 &.936 &0.06 &0.07 &0.15 &0.05 &434 &109 &2.5 &352 &282 &0.44 &0.09\\
3000 &1000 &10 &1.00  &1.00 &1.00 &.987 &.982 &.962 &.983 &.973  &.976  &.971 &.939 &0.23 &0.15 &0.49 &0.21  &705 &315 &6 &1122 &1103 &1.3 &0.18\\
10000 &1000 &10 &1.00  &1.00 &1.00 &.998 &.997 &.972 &.995 &.979  &.979  &.971 &.942 &1.8 &1.8 &1.8 &1.4  &2151 &962 &20 &3789 &3725 &4.9 &0.49\\\hline
1000 \phantom{$^I$}\hspace{-2mm} &1000 &30 &.996  &.995 &.995 &.919 &.923 &.943 &.964 &.954  &.937  &.956 &.936 &0.13 &0.08 &0.29  &0.05  &240 &150 &2.3 &358 &293 &1.2 &0.16\\
3000 &1000 &30 &1.00  &1.00 &1.00 &.969 &.954 &.955 &.984 &.979  &.976  &.975 &.948 &0.26 &0.2 &1.10 &0.29  &565 &395 &6 &1840 &1139 &4.1 &0.48\\
10000 &1000 &30 &1.00  &1.00 &1.00 &.997 &.985 &.965 &.996 &.987  &.984  &.980 &.956 &3.5 &3.3 &3.5 &2.0  &3914 &1265 &20 &3860 &3710 &14 &1.5\\
\hline
\end{tabular}

\end{center}
\vspace{-5mm}
\end{table*}

\begin{table*}[htb]
\small
\begin{center}
\caption{Classification experiments on simulated data with noisy labels, $\delta=0.9$, averaged over 100 runs. 
}\label{tab:noisy}
\vskip -2mm
\begin{tabular}{|l|c|c|c|c|c|c|c|c|c|c|c|c|c|c|c|c|c|c|c|c|c|c|c|c|}
\hline
 \multicolumn{3}{|l|}{} &\multicolumn{11}{|c|}{\hspace{-1mm}\textbf{All-variable detection rate} (\textbf{DR})\hspace{-2mm}} &\multicolumn{11}{|c|}{\phantom{$^I$}\textbf{Percent correctly detected} (\textbf{PCD})}\\
\hline
$N$\phantom{$I^I$}\hspace{-3mm} &$M$ &$\hspace{-.5mm}k\hspace{-1mm}=\hspace{-1mm}k^*\hspace{-1mm}$    &FSA\hspace{-2mm}&FSV\hspace{-1mm} &FSL  &\hspace{-0mm}QTP\hspace{-0mm}  &L1 &EL &L2 &\hspace{-.5mm}MCP\hspace{-.5mm} &\hspace{-0mm}SCD\hspace{-0mm} &\hspace{-0mm}LB\hspace{-0mm} &\hspace{-0.1mm}LB1\hspace{-1mm}  &FSA &FSV\hspace{-2mm}&FSL &QTP   &L1 &EL &L2 &MCP &SCD&LB &LB1\\
\hline
300 \phantom{$^I$}\hspace{-2mm} &1000 &10   & 0 &0 &1 &0   &0 &0 &0 &0 &0  &0 &0  &44.5 &38.9 &43.7  &30.7  &41.2 &35.2 &- &46.7 &45.8 &47.8 &21.8\\
1000\hspace{-2mm} &1000 &10  &45 &45 &86 &0   &0 &0 &0&17  &8  &21  &0 &92.5 &91.4   &98.5  &58.8 &65.3 &44.8 &- &81.2 &78.9 &84.4 &25.4 \\
3000\hspace{-2mm} &1000 &10 &100&100 &100 &20  &22 &0 &0 &66 &58  &91  &0 &100 &100 &100 &88.2   &87.8 &53.9 &- &95.5 &94.2  &99.1 &29.1\\
10000\hspace{-2mm} &1000 &10 &100&100 &100 &100  &92 &2 &0 &95 &95  &100  &0 &100 &100 &100 &100   &99.2 &65 &- &99.5 &99.5  &100 &31.4\\\hline
1000\phantom{$^I$}\hspace{-2mm} &1000 &30 &0 &0 &0 &0  &0 &0 &0 &0 &0  &0  &0 &49.5 &45.0 &53.7  &34.9 &40.0 &35.1 &- &47.5 &47.3 &48.8  &26.7 \\
3000\hspace{-2mm} &1000 &30 &12 &14 &68 &0  &0 &0 &0 &2 &5 &1  &0 &92.4 &92.3 &98.7 &67.5 &63.7 &40.5 &- &84.0 &83.9 &82.8 &32.9\\
10000\hspace{-2mm} &1000 &30 &99 &99 &100 &7  &0 &0 &0 &60 &49 &60  &0 &100 &100 &100 &93.7 &90.3 &47.5 &- &97.5 &96.8 &98.3 &40.7\\
\hline
 \multicolumn{3}{|l|}{} &\multicolumn{11}{|c|}{\hspace{-1mm}\textbf{Area under the ROC curve} (\textbf{AUC})\hspace{-2mm}} &\multicolumn{11}{|c|}{\phantom{$^I$}\textbf{Training Time} (sec)}\\
\hline
$N$\phantom{$I^I$}\hspace{-3mm} &$M$ &$\hspace{-.5mm}k\hspace{-1mm}=\hspace{-1mm}k^*\hspace{-1mm}$    &FSA\hspace{-2mm}&FSV\hspace{-1mm} &FSL  &\hspace{-0mm}QTP\hspace{-0mm}  &L1 &EL &L2 &\hspace{-.5mm}MCP\hspace{-.5mm} &\hspace{-0mm}SCD\hspace{-0mm} &\hspace{-0mm}LB\hspace{-0mm} &\hspace{-0.1mm}LB1\hspace{-1mm}  &FSA &FSV\hspace{-2mm}&FSL &QTP   &L1 &EL &L2 &MCP &SCD&LB &LB1\\
\hline
300\phantom{$^I$}\hspace{-2mm} &1000 &10 &.890  &.868 &.885 &.834 &.880 &.889 &.834 &.877  &.865  &.885  &.863 &0.04 &0.04 &0.04 &0.04   &22 &67 &.44 &37 &65 &0.17 &0.02\\
1000 &1000 &10 &.943  &.940 &.946 &.890 &.907 &.902 &.892 &.914  &.906  &.915 &.888 &0.14 &0.13 &0.14 &0.12 &412  &120 &2 &218&211 &0.53 &0.06\\
3000 &1000 &10 &.950  &.950 &.950 &.935 &.934 &.913 &.928 &.927  &.923  &.924 &.895 &0.49 &0.46 &0.48 &0.36  &1094 &321&8 &385 &367 &1.6 &0.17\\
10000 &1000 &10 &.950  &.950 &.950 &.950 &.949 &.923 &.939 &.933  &.932  &.924 &.897 &2.0 &2.0 &2.1 &1.6  &13921 &940 &20 &653 &599 &5.3 &0.5\\\hline
1000\phantom{$^I$}\hspace{-2mm} &1000 &30 &.905  &.889 &.904 &.845 &.885 &.895 &.895 &.876  &.873  &.898 &.887 &0.30 &0.29 &0.30  &0.17  &791 &142 &1.9 &246 &239 &1.6 &0.17\\
3000 &1000 &30 &.945  &.943 &.949 &.906 &.911 &.907 &.929 &.926  &.919  &.925 &.902 &1.0 &1.0 &1.0 &0.55  &1862 &404&7.9 &522 &504 &4.7 &0.48\\
10000 &1000 &30 &.950  &.950 &.950 &.942 &.938 &.916 &.940 &.937  &.933  &.932 &.908 &3.8 &3.8 &3.8 &2.1  &15949 &1213 &21 &763 &719 &16 &1.6\\
\hline
\end{tabular}

\end{center}
\vspace{-7mm}
\end{table*}

The experiments are performed on the linearly separable data and its noisy version described above.
The algorithms being compared are:
\begin{itemize}
\item FSA - The FSA Algorithm \ref{alg:fsa1} for the logistic loss \eqref{eq:logloss} with  the $\mu=300$ annealing schedule, $\eta=20$.
\item FSV, FSL - The FSA Algorithm \ref{alg:fsa1} for the SVM loss \eqref{eq:svmloss} and Lorenz loss \eqref{eq:lorloss} respectively, with  the $\mu=300$ annealing schedule, $\eta=1$.
\item L1 - The interior point method \cite{koh2007interior} for $L_1$-penalized Logistic Regression using the implementation from {\em http://www.stanford.edu/$\sim$boyd/l1\_logreg/}. To obtain a given number $k$ of variables, the $L_1$ penalty coefficient $\lambda$ is found using the bisection method \cite{burdennumerical}. The bisection procedure calls the interior point training routine about $9$ times until a $\lambda$ is found that gives exactly $k$ nonzero coefficients. Then an unpenalized model was fitted on the selected variables..
\item EL - Elastic net on the Logistic loss with $L_1+L_2$ penalty using the stochastic gradient descent algorithm. We used the Python implementation \verb sklearn.linear_model.SGDClassifier  of \cite{tsuruoka2009stochastic}, 1000 epochs for convergence, and the bisection method for finding the appropriate $L_1$ penalty coefficient. After feature selection, the model was refit on the selected variables with only the $L_2$ penalty $\alpha=0.001$.
\item L2 - SVM using the Python implementation \verb sklearn.linear_model.SGDClassifier  with 1000 epochs, and choosing the $L_2$ penalty coefficient $\alpha\in \{10^{-5},10^{-4},10^{-3}, 10^{-2},10^{-1}\}$ that gave the best result.
\item QTP - The quantile TISP \cite{SheGLMTISP} using a fixed sparsity level with 10 thresholding iterations and 500 more iterations on the selected variables for convergence.

\item MCP, SCD - Logistic regression using MCP (Minimax Concave Penalty)\cite{zhang2010nearly} and  SCAD penalty respectively. Two implementations were evaluated: the \verb ncvreg  R package based on the coordinate descent algorithm \cite{breheny2011coordinate} and the \verb cvplogistic  R package based on the the Majorization-Minimization by Coordinate Descent (MMCD) algorithm \cite{jiang2011majorization}.  The \verb cvplogistic  package obtained better results, which are reported in this paper.
\item LB -  Logitboost using univariate linear regressors as weak learners. In this version, all $M$ learners (one for each variable) are trained at each boosting iteration and the best one is added to the classifier.
\item LB1 - Similar to LB, but only $10\%$ of the learners were randomly selected and trained at each iteration and the best one was added to the classifier.
\end{itemize}

In Tables \ref{tab:l1} and \ref{tab:noisy} are shown the \textit{all-variable} detection rate (DR)\ and the average percent of correctly detected variables, (PCD) obtained from 100 independent runs.
The PCD is the average value of
$|\{j,\beta_j\not =0\}\cap \{j, \beta_j^*\not=0\}|/k^*\cdot 100$.
A more stringent criterion is the
DR which is the percentage of times when all $k^* $ variables were correctly found i.e. $\{j,\beta_j\not =0\}=\{j, \beta_j^*\not=0\}$.
The average area under the ROC curve on unseen data of same size as the training data, and the average training times are also shown in Tables \ref{tab:l1} and \ref{tab:noisy}.

\begin{table*}[t]
\small
\begin{center}
\caption{Regression experiments on simulated  data with correlation $\delta=0.9$, averaged over 100 runs.
}\label{tab:simreg}
\vskip -2mm
\begin{tabular}{|c|c|c|c|c|c|c|c|c|c|c|c|c|c|c|c|c|c|c|}
\hline
 \multicolumn{3}{|c|}{Data Params.} &\multicolumn{8}{|c|}{\textbf{All-variable detection rate} (\textbf{DR})} &\multicolumn{8}{|c|}{\phantom{$^I$}\textbf{Percent correctly detected} (\textbf{PCD})}\\
\hline
$N$\phantom{$^I$} &$M$ &$k$  &FSA  &SA &QTP &L1 &EL &L2 &MCP &SCAD & FSA  &SA & QTP  & L1 &EL &L2  & MCP  &SCAD \\
\hline
300\phantom{$^I$} &1000 &30  &67 &0 &0 &0 &0 &0 &0 &0 &98.5 &- &33.1  &37.8 &37.0 &- &62.1 &24.9\\
1000 &1000 &30  &100 &0 &0 &0 &0 &0 &1 &0 &100 &- &51  &63.8 &62.5 &- &78.5 &55.6\\
3000 &1000 &30  &100 &0 &0  &1 &1 &0 &8 &0 &100 &- &72  &88.6 &87.6 &- &93.4 &83.4\\
10000 &1000 &30  &100 &0 &0 &87 &82 &0 &95 &76 &100 &- &90 &99.5 &99.4 &- &99.8 &99.2\\
\hline
 \multicolumn{3}{|c|}{} &\multicolumn{8}{|c|}{\phantom{$^I$}\textbf{RMSE}}&\multicolumn{8}{|c|}{\textbf{Time}(s)}\\
\hline
$N$\phantom{$^I$} &$M$ &$k$  &FSA &SA  &QTP &L1 &EL &L2 &MCP &SCAD &FSA &SA &QTP &L1 &EL &L2 &MCP &SCAD\\
\hline
300\phantom{$^I$} &1000 &30  &1.11 &16.61 &5.08 &3.68 &3.72 &2.58 &2.41 &6.63 &0.25 &1548 &0.09 &1.24 &1.1 &0.05 &1.56 &0.15\\
1000 &1000 &30  &1.02 &2.26 &4.22 &2.79 &2.88 &2.45 &3.17 &6.51 &0.18 &2730 &0.1 &2.6 &2.3 &0.07 &3.5 &7.6\\
3000 &1000 &30  &1.01 &1.64 &3.06 &1.87 &1.94 &1.22 &3.80 &6.36 &0.52 &10251 &0.3 &5.7 &5.6 &0.16 &12 &41\\
10000 &1000 &30  &1.00 &1.51 &1.91 &1.05 &1.07 &1.05 &3.94 &5.01 &1.8 &37296 &1.0 &14 &14 &0.33 &41 &79\\
\hline
\end{tabular}

\end{center}
\vspace{-7mm}
\end{table*}

The FSA algorithm detects the true variables more often and obtains significantly better ($p$-value $<10^{-4}$) AUC numbers than the other algorithms. At the same time the training time is reduced by three orders of magnitude compared to the penalized methods and is on par with TISP and Logitboost.

The $L_1$ penalized logistic regression needs about ten times more data to obtain a similar performance as the FSA. On the noisy data, the MCP and SCAD methods cannot always reach the 5\% Bayes error, even for large data. This is probably because they sometimes get stuck in a weak local optimum. The elastic net (EL) based on stochastic gradient descent  is behind in terms of variable selection, and is competitive in terms of  prediction only for small data sizes. The $L_2$ penalized SVM does a good job at prediction for large data sizes, but the FSA can do a better job faster and using 3-10 times less data.

We observe that given sufficient training data, the FSA algorithm will always find the true variables and learn a model that has almost perfect prediction on the test data. These findings are in accord with the theoretical guarantees of convergence and consistency from Theorem \ref{thm:consist}.


\vspace{-2mm}
\subsubsection{Large (Correlated) Data Experiment } \label{sec:bigdata}
\vspace{-0mm}

 In this subsection we experiment with even larger datasets with equally correlated predictors. The equally correlated case is well known to be challenging in the feature selection literature.  Generating such big datasets is not trivial (the conventional covariance matrix based simulation could easily run out of memory). One of our computationally efficient  ways delivers  observations $\bx_i$  as follows: first generate $z_i\overset{iid}{\sim} N(0, 1)$, then set
\vspace{-2mm}
\begin{equation}
\bx_i = \alpha z_i \b1_{M\times 1} + \be_i,  \mbox{ with } \be_i\sim N(0,I_M),  \label{eq:bigdata}
\vspace{-2mm}
\end{equation}
 and obtain $\bsbX=[\bx_1^T, \cdots, \bx_M]^T$. It is easy to verify that for the correlation between \textit{any} pair of predictors is $\alpha^2/(1+\alpha^2)$.  We set $\alpha=0.5$. 

The algorithms being compared are FSA with Lorenz loss and Logitboost, named FSL and LB1 in   Section \ref{sec:expl1}.
In Figure \ref{fig:bigdata} are shown the training times (left), percent of variables correctly detected (middle) and area under ROC curve (right) for $N\leq 10^6, M=10^5, k=k^*=500$, averaged over 10 runs.

Recall that Logitboost in implementation works with the columns of the data matrix (where the observations are stored as rows). Thus  Logitboost   is severely limited by the memory capacity and could  handle up to 70,000 observations  in the experiments. Although data reloading  appears to be  an option,  with almost \textit{all} variables needed to be reloaded at each boosting iteration, the computation is  very slow for say  $k=500$ iterations. In contrast,  in FSA one can re-load  only the working variables, the number of which quickly drops to an affordable number (and keeps decreasing as the algorithm proceeds). Starting with $100,000$ features, after less than $20$ iterations the working features for all $1,000,000$ observations can be stored in memory in our experiments.

According to Figure \ref{fig:bigdata},  one could see that FSA is at least 10 times faster than Logitboost and can handle much larger datasets. At the same time, it is better at detecting the true variables and has better prediction than Logitboost.

\vspace{-2mm}
\subsubsection{Regression Experiments}\label{sec:reg}

Similar to the classification simulations, the observations are sampled from a multivariate normal $\bx \sim {\cal N}(0,\Sigma )$ where $\Sigma_{ij}=\delta^{|i-j|}$ and $\delta=0.9$. Given $\bx$, the dependent variable $y$ is obtained as
\vspace{-2mm}
\[
y=\sum_{i=1}^{k^*}x_{10i}+\epsilon, \; \epsilon \sim  {\cal N}(0,1)
\vspace{-2mm}
\]
We experimented with different data sizes and number $k^*$ of relevant variables. The results of the experiments, averaged over 100 runs, are given in Table \ref{tab:simreg}.

The following algorithms were evaluated:
\begin{enumerate}
\item FSA - The FSA Algorithm \ref{alg:fsa1} with  the $\mu=300$ annealing schedule, $\eta=20$.
\item SA - Feature Selection by Simulated Annealing, a wrapper method on OLS linear regression that uses simulated annealing to select the best variable set by 2-fold cross-validation on the training set. It is implemented using the \verb caret  R package, 200 iterations and 10 restarts.
\item QTP - The quantile TISP algorithm with 10 thresholding iterations and 500 more iterations on the selected variables for convergence.
\item L1 - The built in \verb lasso  function from Matlab. The model was refit on the selected variables by least squares.
\item EL - Elastic net with the built in \verb lasso  function from Matlab with mixing coefficient $0.99$. The model was refit on the selected variables by least squares with shrinkage penalty $0.01$.
\item L2 - OLS linear regression with the shrinkage ($L_2$ penalty)  $\alpha\in \{1,0.1,0.01,10^{-3},10^{-4},10^{-5}\}$ that gave the best result.
\item MCP, SCD.  The MCP and SCAD penalized regression using coordinate descent  \cite{breheny2011coordinate}. The \verb ncvreg  C++ implementation was used.
\end{enumerate}

One could see from Table \ref{tab:simreg} that the FSA algorithm consistently finds the true variables more often than the other methods and obtains better predictions in terms of root mean square error (RMSE) than the other methods. The other methods need at least ten times more data to obtain a similar performance to the FSA method. The simulated annealing based method does a decent job but is extremely slow.

We also observe that the FSA algorithm scales quite well to large data sizes, in fact it scales as $O(MN)$ where $N$ is the number of observations and $M$ is the number of variables.

\vspace{-3mm}
\subsection{UCI Data Experiments} \label{sec:uci}

We applied the FSA algorithm to three datasets from the UCI Machine Learning Repository:
the URL reputation \cite{ma2009identifying} (large size), Gisette  (medium size) and Dexter (small).
The FSA in these experiments used the Lorenz loss and $\mu=300,\eta=1,N^{iter}=500$.

The URL\_Reputation data is about classifying websites into malicious/non-malicious, based on a feature vector of size up to 3.2 million. The data is organized in days, with 20,000 observations in each day. The training set contains the first 100 days, and the test set is the 101-th day. On the URL\_Reputation dataset we compared with the confidence-weighted online learning algorithm \cite{ma2009identifying}, SVM and Logistic Regression with Stochastic Gradient descent, as reported by  \cite{ma2009identifying}.
We see from Table \ref{tab:url} that the Linear FSA with the Lorenz loss obtains a test error of $1.15\%$ by selecting 75,000 features, close to the Confidence-Weighted algorithm error of $1.0\%$. The errors of SVM and Logistic Regression with Stochastic Gradient Descent, which did not perform any feature selection, were $1.8\%$ and respectively $1.6\%$.

The other two datasets are part of the Feature Selection Challenge 2003, and the comparisons were obtained from the challenge website. Exception is the Feature Selection by Simulated Annealing (SA) entry, described in Section \ref{sec:reg}, using the same parameters and linear SVM. Note that the FS Challenge website is now down and test results could not be obtained for SA.

On the Gisette dataset, the Linear FSA algorithm does a decent job compared to the Recursive Feature Elimination (RFE) and the MCP-penalized method (MCP), and better than the Parallel FS \cite{zhou2014parallel} and SA. Of course there are many methods that perform even better, using wrapper methods, non-linear mappings line neural networks, some of them not using feature selection at all.

\begin{table}[htb]
\vspace{-4mm}
\normalsize
\begin{center}
\caption{UCI data experiments.
}\label{tab:url}
\vskip -2mm
\begin{tabular}{lcccc}
\hline
&Number of &Error &Error &Error \\
Method  &features $k$ &train \% &valid \%&test \%\\
\hline
 \multicolumn{5}{l}{URL\_Reputation, $M$=3.2$\cdot \hspace{-0.2mm}10^{6\phantom{^I}}\hspace{-1.5mm}$,$N^{train}$=2$\cdot \hspace{-0.2mm}10^6\hspace{-0.5mm}$,$N^{test}$=20000} \\
\hline
SVM    &all &- &-&1.8\\
Log Reg-SGD  &all &-&-&1.6\\
FSA   &75,000 &0.50 &-&1.15\\
CW \cite{ma2009identifying}    &500,000 &0.23 &-&1.0\\
\hline
 \multicolumn{5}{l}{Gisette, $M$=5000,$N^{train\phantom{^I}}\hspace{-1.5mm}$=6000,$N^{valid}$=1000,$N^{test}$=6500} \\
\hline
RFE &700 &0.38 &1.2 &1.82\\
FSA*  &500 &0.2 &0.1 &1.83\\
MCP &1185 &0.88 &1.5 &1.85\\
FSA, 2 bins   &320 &1.13 &1.3 &1.88\\
FSA, 5 bins   &200 &0.67 &1.4 &1.95\\
Parallel FS \cite{zhou2014parallel} &500 &- &- &2.15\\
SA &2258 &0. &4.1 &-\\
\hline
\multicolumn{5}{l}{Dexter, $M$=20000,$N^{train\phantom{^I}}\hspace{-1.5mm}$=300,$N^{valid}$=300,$N^{test}$=2000} \\
\hline
FSA*  &300 &0. &0. &8.55\\
FSA*  &93 &2.0 &1.0 &8.7\\
FSA  &93 &0.33 &8.67 &10.3\\
L1 \cite{rosset2003piecewise} &93 &0.33 &9.0 &6.3\\
SA &8054 &0. &12.0 &-\\
\hline
\end{tabular}
\end{center}
\vspace{-6mm}
\end{table}
On the Dexter data with 300 training observations and 20000 features, the linear FSA overfits. The best test error of $8.55$ was achieved by training on both training and validation sets. Again, the data being small, many methods including wrapper methods and nonlinear mappings can be used to do a better job than a linear classifier. The training times for FSA and SA are shown in Table \ref{tab:times}, showing that SA is much slower than FSA.
\begin{table}[htb]
\vspace{-3mm}
\normalsize
\begin{center}
\caption{Training times in seconds.
}\label{tab:times}
\vskip -2mm
\begin{tabular}{lccc}
Dataset &SA&FSA &FSA w/bins \\
\hline
Gisette  &144,655&4.0 &12 \\
Dexter &9,038&0.5 &1.2 \\
\hline
\end{tabular}
\end{center}
\vspace{-9mm}
\end{table}
\subsection{Face Keypoint Detection Experiments} \label{sec:detclf}

As this feature selection method is intended to be used in computer vision, we present experiments on detecting face keypoints from color images. The face keypoints such as eye centers, nose sides, mouth corners, chin, bottom of ears, are represented as 2D points $(x,y)$.


{\bf AFLW.}  The dataset used for training and testing is the AFLW dataset \cite{koestinger11b}, which has 21123 images containing 24386 faces annotated with 21 points. Of them, 16207 images were found to contain one face per image and 999 of them were selected for training (AFLWT). There were 2164 images containing at least 2 annotated faces. By visual inspection, 1555 of them were found to have all the faces annotated and were used as the test dataset AFLWMF. These 1555 images contain 3861 faces.

{\bf Feature pool.} All classifiers were trained using a feature pool consisting of $288\times 3=864$ Histograms of Oriented Gradients (HOG) features \cite{dalal2005histograms} and 61000 Haar features extracted from the RGB channels in a $24\times 24$ pixel window centered at the point of interest $(x,y)$ in one of the images of a Gaussian pyramid with 4 scales per octave (i.e. resized by powers of $2^{1/4}$).
\begin{figure*}[ht]
\centering
\vspace{-1mm}
\includegraphics[height=4.1cm]{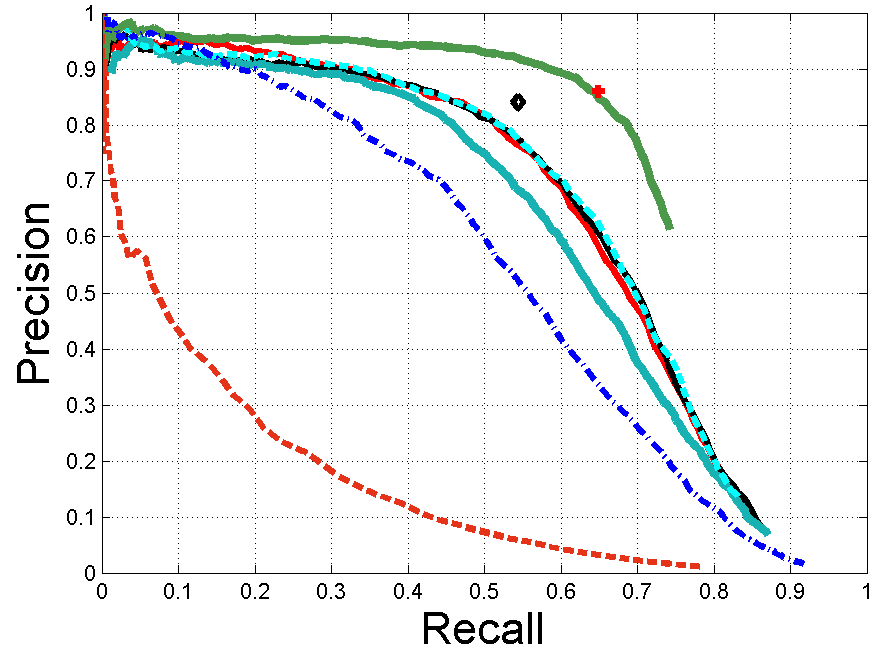}
\includegraphics[height=4.1cm]{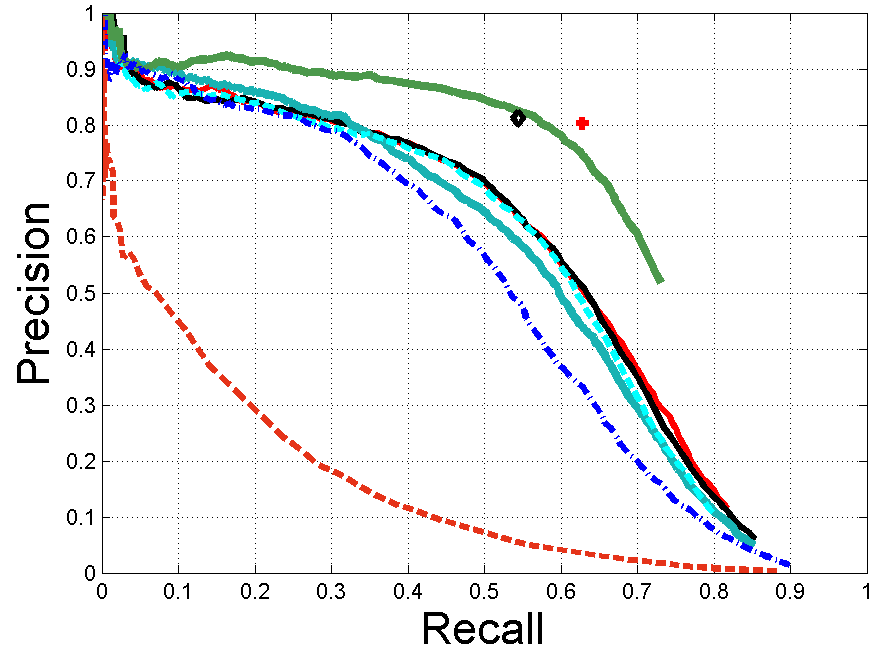}
\includegraphics[height=4.1cm]{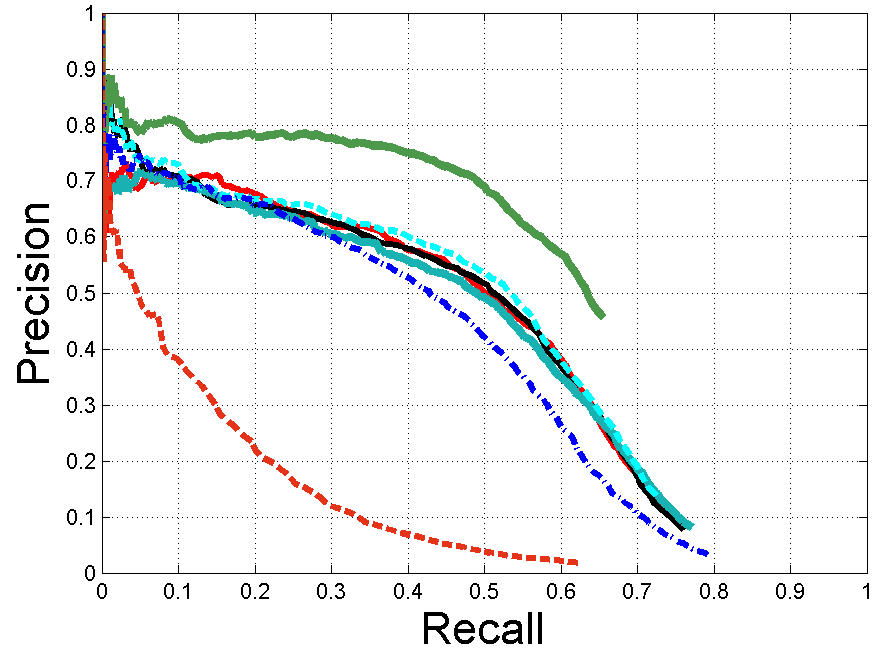}
\includegraphics[height=4.1cm]{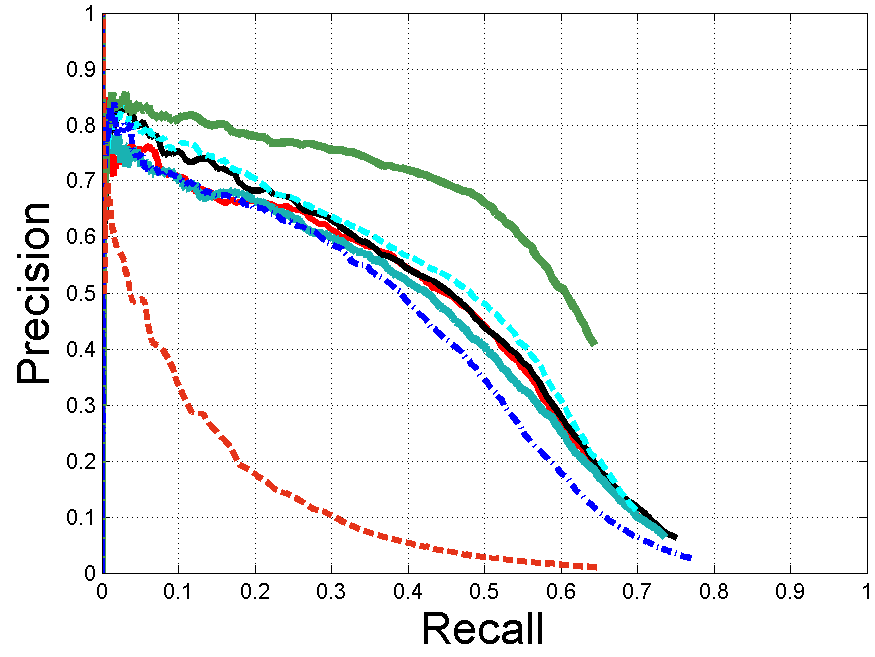}
\includegraphics[height=4.1cm]{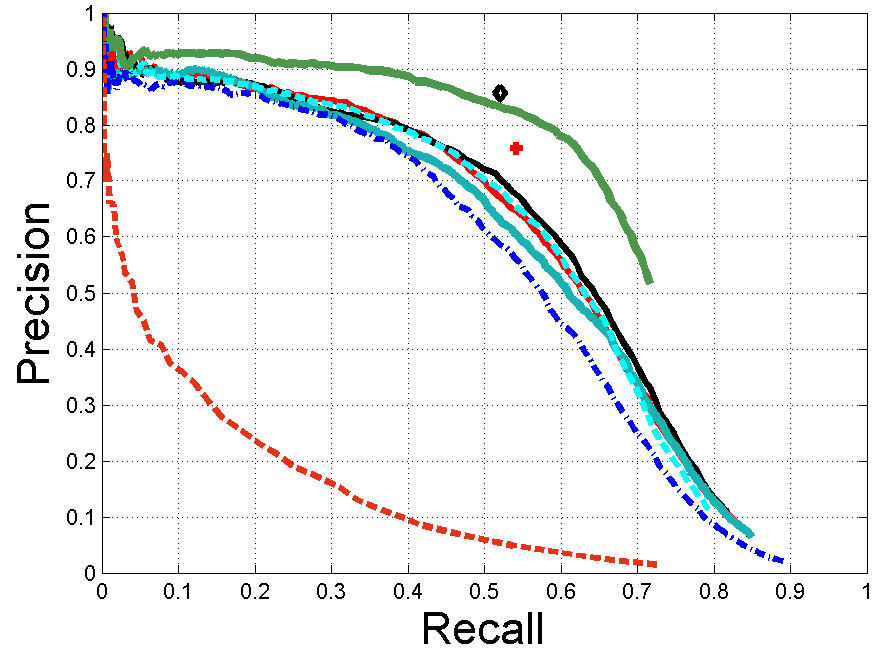}
\includegraphics[height=4.1cm]{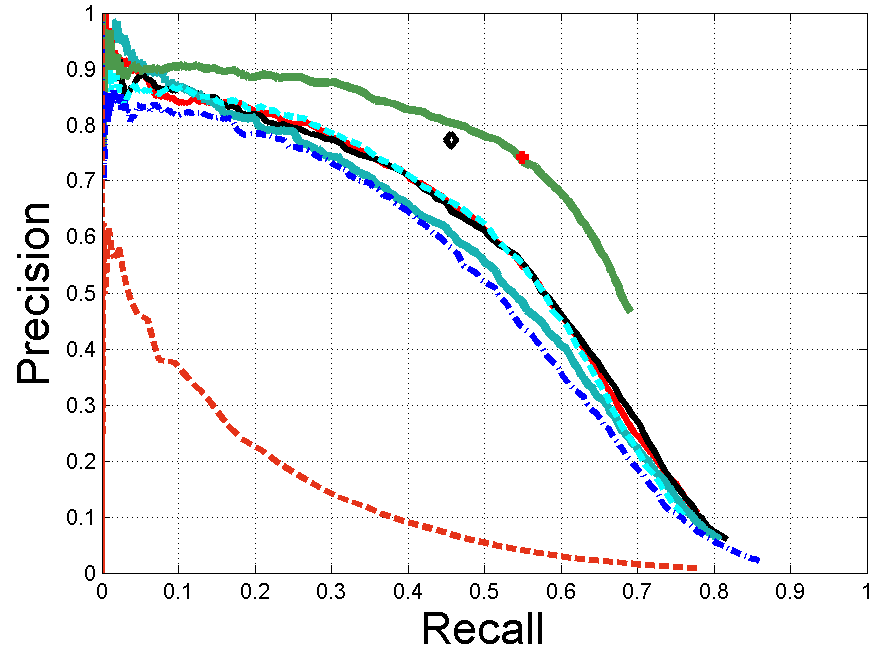}
\includegraphics[height=4.1cm]{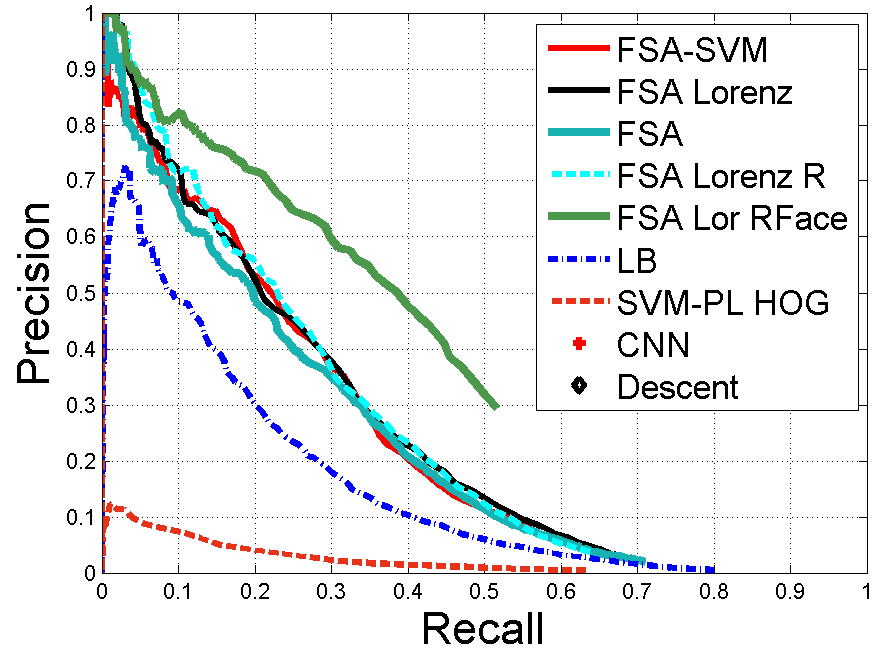}
\includegraphics[height=4.1cm]{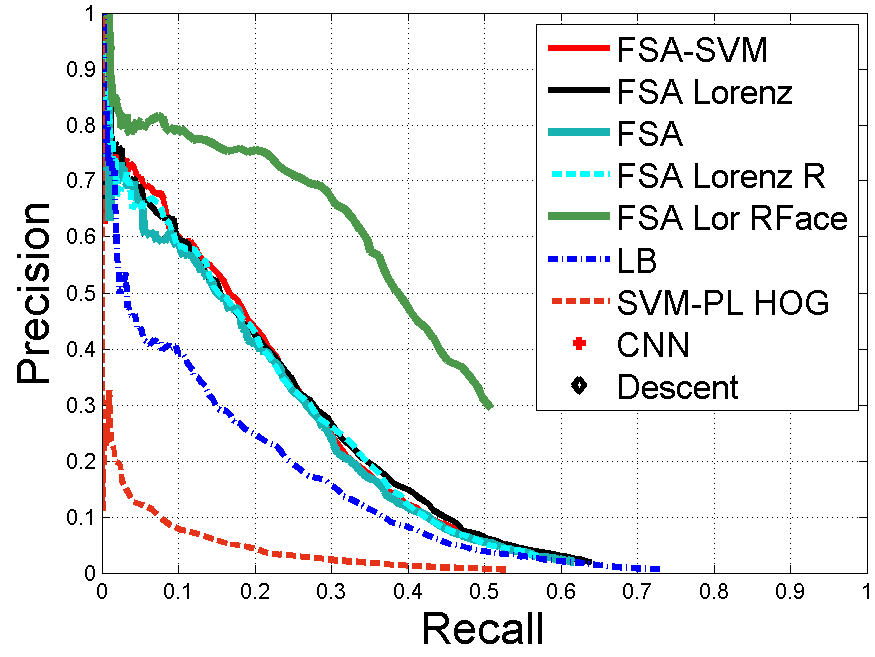}
\includegraphics[height=4.1cm]{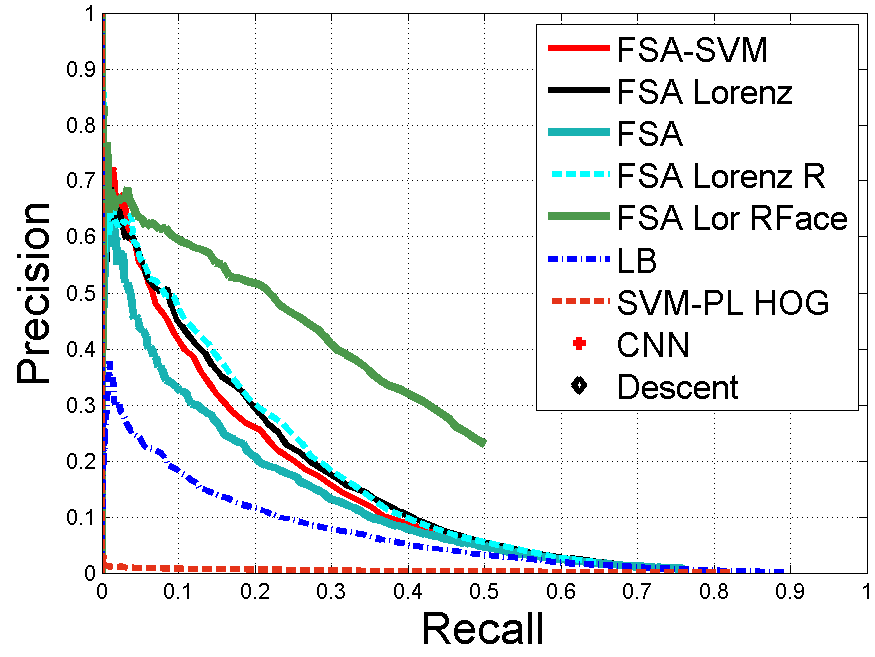}
\vskip -3mm
\caption{Precision-recall curves for face keypoint detection on the test set AFLWMF containing 1555 images and 3861 faces. From top to bottom, left to right: left/right eye center, left/right nose, left/right mouth corner, left/right ear, chin.}
\label{fig:pr_curves}
\vspace{-6mm}
\end{figure*}

{\bf Training examples.} The training examples are points on the Gaussian pyramid, with the positives within one pixel from the keypoint annotation on the images of the pyramid where the inter-eye distance (computed by fitting a rigid 3D face model)  is in the $[20,40]$ pixel range. The negatives are all points at least 0.5 IED (inter-eye distance) from the keypoint annotation. In total the 999 AFLWT training images contain about 1 billion negatives. All the negatives were used for training the classifiers through a negative mining procedure similar to \cite{felzenszwalb2010object}, with the difference that about 20,000 hard negatives were added to the training set at each iteration, thus the set of training negatives increased with each mining iteration. All classifiers were trained with 10 iterations of mining hard negatives.

{\bf Classifier size.} A separate classifier was trained for each keypoint being evaluated. All classifiers except SVM-PL HOG were trained as monolithic classifier with 1500 features or weak learners. The SVM-PL HOG classifier was trained on all 864 HOG features, without feature selection

{\bf Detection criteria.} The following criteria were used for evaluating detection performance. The visible face keypoint is considered detected in an image if  a detection is found at most 5\% of the IED away in one of the images of the pyramid. A detected point $p$ in one of the images of the pyramid is a false positive if it is at least 10\% of the IED away from the face part being evaluated (visible or not) of any face of the image.

\vspace{-2mm}
\subsubsection{Results}

{\bf Algorithms.} We compared the following learning algorithms:
\begin{enumerate}
\item FSA-Logistic, FSA-SVM, FSA-Lorenz  - The FSA method on the Logistic \eqref{eq:logloss}, SVM \eqref{eq:svmloss}, and Lorenz losses \eqref{eq:lorloss} with piecewise linear learners, $\mu=300, N^{iter}=500$.
\item LogitBoost using univariate piecewise constant regressors as weak learners. For speed reasons, only $10\%$ of the learners were selected at random and trained at each boosting iteration and the best one was added to the classifier.
\item SVM-PL HOG - The SVM algorithm with piecewise linear response on each variable. The variables were the 864 HOG features.
\end{enumerate}

 In Figure \ref{fig:pr_curves} are shown the  precision-recall curves for detecting nine keypoints on the AFLWMF data. 
One can see that the FSA-SVM and FSA-Lorenz perform similarly and slightly outperform the FSA on the logistic loss. All three FSA versions outperform Logitboost and greatly outperform the piecewise linear SVM on the HOG features.
At the same time, training the FSA algorithm is about 8 times faster than the LB algorithm, which is 10 times faster than the full LB version that trains all weak learners at each boosting iteration. 

Also shown are the supervised descent method \cite{xiong2013supervised}  and the CNN based face point detection method \cite{sun2013deep} on the eye and mouth (the keypoints that were in common with the keypoints that we evaluated).

These two methods outperform the classification and regression-based FSA detectors. However, we must point out that the two face alignment methods are top-down methods that rely on the face being detected first by a face detector, which in the case of the CNN method was trained with about 100k faces. In contrast, our point detectors are bottom-up detectors that were trained with 999 faces to directly detect the keypoints without the intermediary step of finding the face. If we involve our own 3D-model based face detector \cite{barbu2014face} that uses all nine FSA-Lorenz keypoint detectors to detect the face and its 3D pose, we obtain the curve denoted as FSA-Lor Face. These results were obtained using a top-down pruning step that keeps only the keypoint detections that are within 0.5 IED (Inter-Eye Distance) from the predicted locations from the 3D pose. We see that using the top-down information we obtain results comparable to the CNN method \cite{sun2013deep} and slightly better than the supervised descent method \cite{xiong2013supervised}.


\vspace{-2mm}
\subsection{Ranking Experiments}

Sparse motion segmentation is the problem of grouping a given set of trajectories of feature points (that were tracked through the frames of an image sequence) into a number of groups according to their common, usually rigid, motion.
A popular method for sparse motion segmentation is spectral clustering \cite{lauer2009spectral}, where the feature point trajectories are projected to a lower dimensional space where spectral clustering is performed according  to an affinity measure.

The FSA for Ranking using the loss \eqref{eq:rankloss} and piecewise linear response functions was used for ranking a number of candidate motion segmentations obtained by spectral clustering with different parameters to predict the best one. For each segmentation about 2000 features are extracted, measuring model fitness and cluster compactess. Details about how the candidate segmentations are generated and the feature pool are given in the supplementary material.
\begin{figure*}[htb]
\centering
  \includegraphics[height=3.8cm]{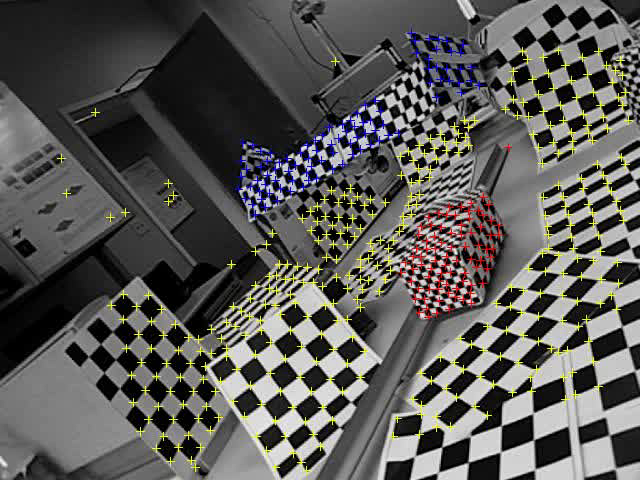}
  \includegraphics[height=3.8cm]{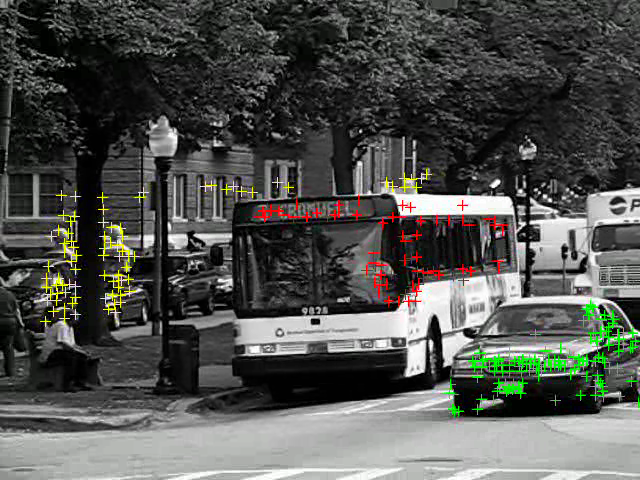}
  \includegraphics[height=3.8cm]{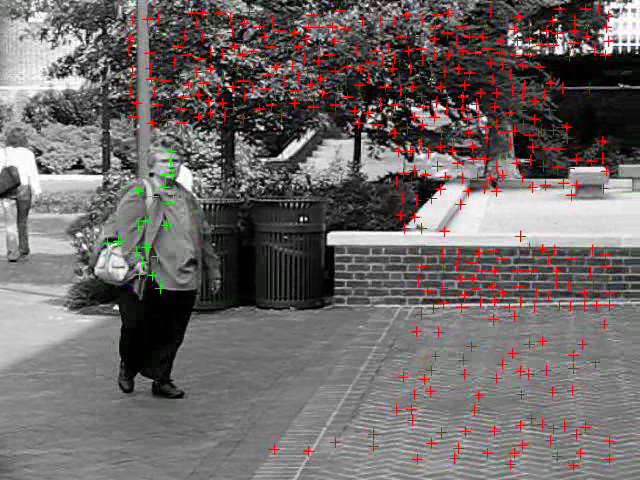}
\vskip -2mm
\caption{Sample images from some sequences of three categories in the Hopkins
  155 database with the ground truth labels superimposed.}
\label{fig:hopkins155}
\vspace{-5mm}
\end{figure*}

The FSA Rank based method for motion segmentation was evaluated on the Hopkins 155 dataset~\cite{tron2007benchmark}.
The Hopkins 155 Dataset contains 155 sets of feature point trajectories of 2 or 3 motions from 50 videos, along with the corresponding ground truth segmentation. Based on the content of the video and the type of motion, the 155 sequences can be categorized into three main groups: {\it checkerboard}, {\it traffic} and {\it articulated}. Figure~\ref{fig:hopkins155} shows sample frames from three videos of the Hopkins 155 database with the feature points superimposed.
\begin{table*}[htb]
\small
\begin{center}
  \caption{Misclassification rate (in percent) for sequences of full
    trajectories in the Hopkins 155 dataset.}
\label{tab:hopkins155-result}
\vskip -2mm
\begin{tabular}{|c|c|c|c|c|c|c|c|c|c|c|c|c|}
  \hline
  Method\phantom{$^I$}  &RV & SC & SSC & VC  & \multicolumn{2}{|c|}{RankBoost} & \multicolumn{6}{|c|}{FSARank}\\
\cline{8-13}\vspace{-3.4mm}\\
&&&& &  \multicolumn{2}{|c|}{} & \multicolumn{2}{|c|}{Likelihood Features} & \multicolumn{2}{|c|}{Prior Features} &\multicolumn{2}{|c|}{All\phantom{$^I$} Features}\\
&&&&& Train &Test &Train &Test & Train & Test & Train & Test \\
      \hline
      \hline
      \multicolumn{7}{l}{Checkerboard (2 motion)} \\
      \hline
      Average\phantom{$^I$} &- & 0.85 & 1.12 & 0.67 & 0.67 & 0.74 & 0.58 & 0.69 & 1.09 & 1.28 & 0.12 & 0.12 \\
      Median &- & 0.00 & 0.00 & 0.00 & 0.00  & 0.00 & 0.00 & 0.00 & 0.00 & 0.00 & 0.00 & 0.00 \\
      \hline
      \multicolumn{7}{l}{Traffic (2 motion)} \\
      \hline
      Average\phantom{$^I$} &-& 0.90 & 0.02 & 0.99 & 0.69 & 0.72 & 0.80 & 0.76 & 4.25 & 4.25 & 0.59 & 0.58 \\
      Median  &-& 0.00 & 0.00 & 0.22 & 0.00 & 0.00 & 0.15 & 0.00 & 0.00 & 0.00 & 0.00 & 0.00 \\
      \hline
      \multicolumn{7}{l}{Articulated (2 motion)} \\
      \hline
      Average\phantom{$^I$} &-& 1.71 & 0.62 & 2.94 & 2.05 & 2.26 & 2.30 & 2.27 & 1.32 & 1.32 & 1.32 & 1.32 \\
      Median  &-& 0.00 & 0.00 & 0.88 & 0.00 & 0.00 & 0.00 & 0.00 & 0.00 & 0.00 & 0.00 & 0.00 \\
      \hline
      \multicolumn{7}{l}{All (2 motion)} \\
      \hline
      Average\phantom{$^I$} &0.44 & 0.94 & 0.82 & 0.96 & 0.80 & 0.87 & 0.80 & 0.85 & 1.93 & 2.05 & 0.35 & 0.35 \\
      Median  &-& 0.00 & 0.00 & 0.00 & 0.00 & 0.00 & 0.00 & 0.00 & 0.00 & 0.00 & 0.00 & 0.00 \\
      \hline
      \hline
      \multicolumn{7}{l}{Checkerboard (3 motion)} \\
      \hline
      Average\phantom{$^I$} &-& 2.15 & 2.97 & 0.74 & 0.85 & 2.60 & 0.74 & 0.74 & 4.10 & 4.22 & 0.49 & 0.49 \\
      Median  &-& 0.47 & 0.27 & 0.21 & 0.26 & 0.26 & 0.21 & 0.21 & 0.24 & 0.24 & 0.21 & 0.21 \\
      \hline
      \multicolumn{7}{l}{Traffic (3 motion)} \\
      \hline
      Average\phantom{$^I$} &-& 1.35 & 0.58 & 1.13 & 4.15 & 4.24 & 1.13 & 1.13 & 4.05 & 4.05 & 1.73 & 1.07 \\
      Median  &-& 0.19 & 0.00 & 0.21 & 0.00 & 0.47 & 0.00 & 0.00 & 0.00 & 0.00 & 0.00 & 0.00 \\
      \hline
      \multicolumn{7}{l}{Articulated (3 motion)} \\
      \hline
      Average &-& 4.26 & 1.42 & 5.65 & 3.66 & 18.09 & 5.32 & 5.32 & 3.19 & 3.19 & 3.19 & 3.19 \\
      Median  &-& 4.26 & 0.00 & 5.65 & 3.66 & 18.09 & 5.32 & 5.32 & 3.19 & 3.19 & 3.19 & 3.19 \\
      \hline
      \multicolumn{7}{l}{All (3 motion)} \\
      \hline
      Average &1.88 & 2.11 & 2.45 & 1.10 & 1.67 & 3.82 & 1.08 & 1.08 & 4.04 & 4.13 & 0.90 & 0.76 \\
      Median  &- & 0.37 & 0.20 &0.22 & 0.20 & 0.32 & 0.20 & 0.20 & 0.20 & 0.20 & 0.00 & 0.00 \\
      \hline
      \hline
      \multicolumn{7}{l}{All sequences combined} \\
      \hline
      Average &0.77 & 1.20 & 1.24 & 0.99 & 1.00 & 1.54 & 0.86 & 0.90 & 2.40 & 2.52 & 0.47 & 0.44 \\
      Median  &-& 0.00 & 0.00 & 0.00 & 0.00 & 0.00 & 0.00 & 0.00 & 0.00 & 0.00 & 0.00 & 0.00 \\
      \hline
\end{tabular}
\end{center}
\vspace{-8mm}
\end{table*}

\noindent{\bf Ten Fold Cross Validation.} The Hopkins 155 dataset contains sequences from 50 videos. The 50 videos were divided at random into 10 subsets, each subset containing 5 videos. The 155 Hopkins sequences were also divided into 10 subsets, each subset containing all sequences corresponding to one of the 10 subsets of 5 videos. The reason for separating the videos first and then the sequences is fairness.
Some 2 motion sequences are subsets of 3 motion sequences, and it is possible that the segmentation from 2 motions is a subset of that of 3 motions. If this happens, then it would be unfair to have a 3-motion sequence in the training set and a 2-motion subset from the same sequence for testing.

At round $k$ of the cross validation, we select the $k-$ th of the 10 subsets of sequences as the test set and form the training set from the remaining 9 subsets. After training, we apply the obtained ranking function to rank the motion segmentations for each sequence. The best ranked one is picked as the
final result to calculate the misclassification rate.

\noindent{\bf Parameter Settings.} The parameters for our FSA-Rank method were: number of bins $B=4$, the number of selected features $k=40$.
The other parameters are $N^{iter}=300, \eta=0.5, \mu=300, \lambda=0.01$.

We compared the FSA-Rank method with the RankBoost \cite{freund2003efficient} algorithm  based on the same features and decision stumps as weak rankers (as described in the supplementary material) with the following parameters: the number of thresholds for the decision stumps $B=64$, and the number of boosting iterations was set to 100.

\vspace{-1mm}
\subsubsection{Misclassification Error}
\noindent{\bf Ranking Accuracy.} In Table \ref{tab:hopkins155-result} are shown the average
misclassification errors over all sequences when they were in the training set and when they were in the test set.  Other methods are compared, such as randomized voting (RV) \cite{jungrigid}, spectral clustering (SC) \cite{lauer2009spectral}, sparse spectral clustering (SSC) \cite{elhamifar2009sparse} and velocity clustering (VC) \cite{ding2012motion}.
\begin{figure*}[htb]
\vspace{-2mm}
\centering
\includegraphics[width=7.3cm]{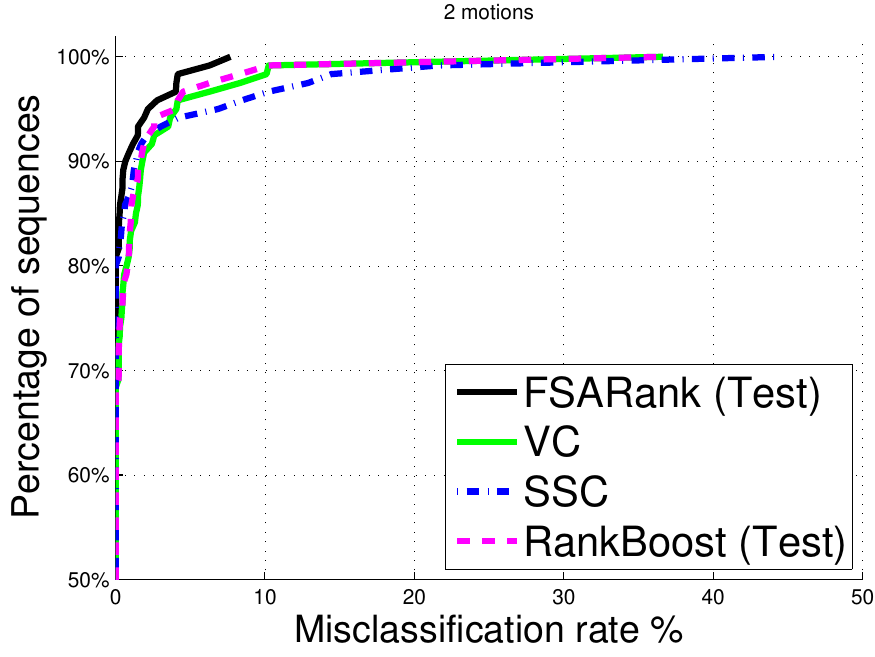}
\includegraphics[width=7.3cm]{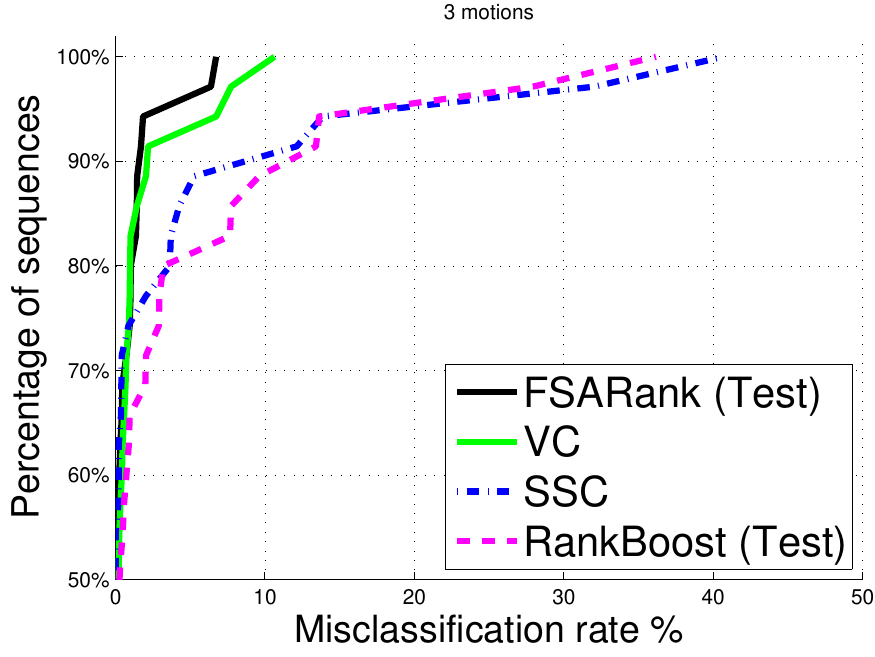}
\vskip -3mm
\caption{The cumulative distribution of the  misclassification rate for two
  and three motions in the Hopkins 155 database. }
\label{fig:cumulative-dist}
\vspace{-6mm}
\end{figure*}

Our method outperforms
 RankBoost in every category on both training and test sets, even though Rankboost uses 100 boosting iterations (thus about 100 features) while FSA-Rank uses only 40 features.

Also the difference in misclassification rate between the training set and test set is very small for FSA-Rank, especially for 2-motion sequences.
In comparison, the average misclassification rate of 3 motions on test set of RankBoost is about 50\% larger than that on training set, while
these two misclassification rates are quite close on our method. This is probably due to the small number of features selected and the shrinkage prior \eqref{eq:prior2}, which together helped obtain a small training error and good generalization.

Compared to VC~\cite{ding2012motion} which uses a fixed measure to select best segmentation, our
method works better on all categories. Moreover, the average
misclassification rates of our method on both 2 motions and 3 motions are
almost half of those from SC~\cite{lauer2009spectral}.

From the cumulative distributions shown in Figure~\ref{fig:cumulative-dist}, we see
that for 2 motions our method performs much better than the other methods compared, while for 3 motions our method is comparable to the best (VC). Nevertheless, our method outperforms RankBoost in both situations.

\vspace{-2mm}
\section{Conclusion and Future Work}

This paper presented a novel learning scheme for feature selection in high dimensional data applications.  It gradually identifies and removes some irrelevant  variables and proceeds according to  an annealing  schedule. We showed that it  solves a constrained optimization problem and has  a performance guarantee  in both  estimation and selection.

As opposed to  the $L_1$ penalized method, the proposed method runs much more efficiently and does not introduce any undesired bias in estimation. It kills variables progressively based on  their importance, which is opposite to the model growing process  of boosting, but  usually brings improvement in variable selection and prediction.

The algorithm  is suitable for big data computation due to its simplicity and ability to reduce the problem size throughout the iteration. In contrast to boosting, the total amount of data the algorithm needs to access for training is only about \textbf{2-10} times the size of the training set, which   makes it  amenable for large scale problems.  Hence in computation,  FSA has similar  advantages as an online algorithm (that accesses each training observation  once) while being much more accurate.  Our approach applies generically to many types of problems, including regression, classification and ranking for instance.
Extensive  experiments on both synthetic data and real data support FSA as a competitive alternative to many up-to-date  feature selection methods.

In the future we plan to apply the variable selection method to challenging object detection problems.
\vspace{-2mm}
{\small
\bibliographystyle{ieee}
\bibliography{featselectbib}
}
\vspace{-13mm}
\begin{IEEEbiography}
[{\includegraphics[width=1in,height=1.25in,clip,keepaspectratio]{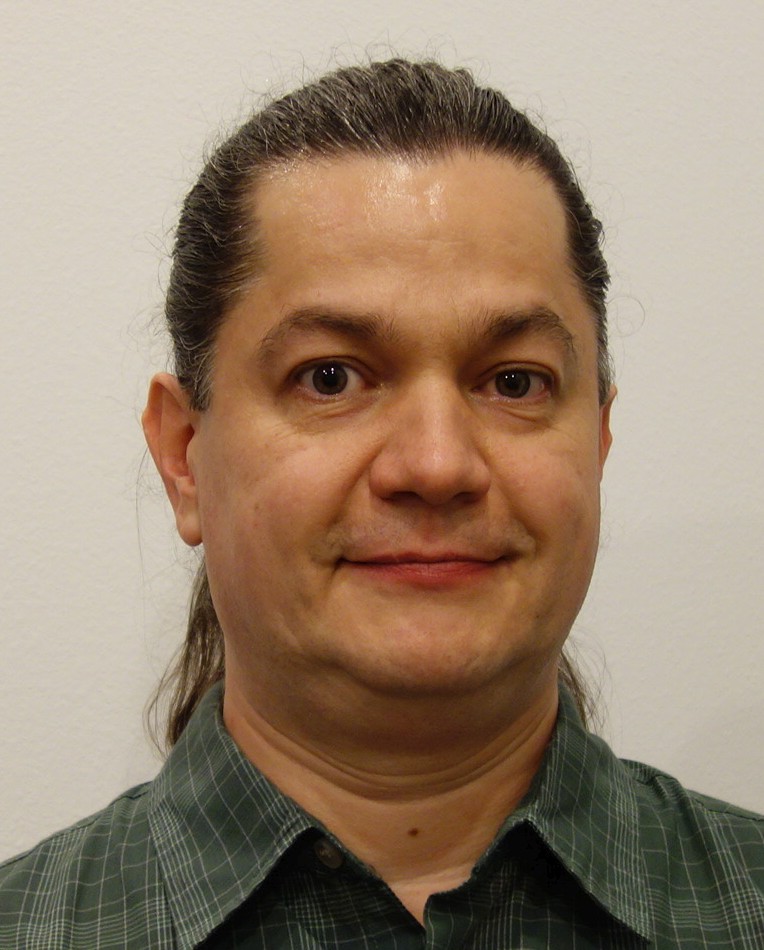}}]{Adrian Barbu}
Adrian Barbu received his BS degree from University of
Bucharest, Romania, in 1995, a Ph.D. in Mathematics from Ohio State University in 2000 and a Ph.D. in Computer Science from UCLA in 2005.
From 2005 to 2007 he was a research scientist and later project manager in Siemens Corporate Research, working in medical imaging.
He received the 2011 Thomas A. Edison Patent Award with his co-authors for their work on Marginal Space Learning.
From 2007 he joined the Statistics department at Florida State University, first as assistant professor, and since 2013 as associate professor.
 His research interests are in computer vision, machine learning and medical imaging.
\vspace{-12mm}
\end{IEEEbiography}

\begin{IEEEbiography}[{\includegraphics[width=1in,height=1.25in,clip,keepaspectratio]{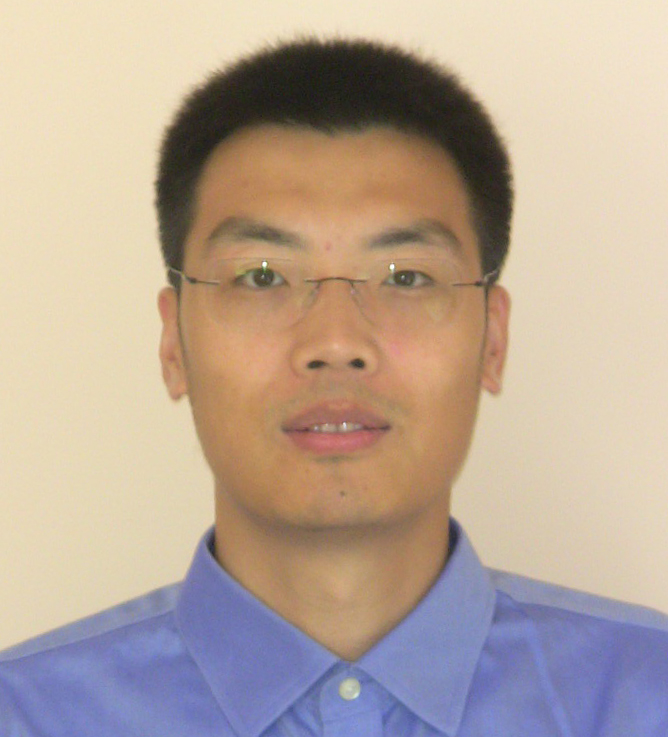}}]{Yiyuan She}
Yiyuan She received the B.S. degree in Mathematics and the M.S.
degree in Computer Science from Peking University in
2000 and 2003, respectively, and received the Ph.D. degree in
Statistics from Stanford University in 2008. He is
currently an associate professor in the Department of
Statistics at Florida State University. His research interests
include high-dimensional statistics, machine
learning, robust statistics, statistics computing,
bioinformatics, and network science.
\vspace{-12mm}
\end{IEEEbiography}

\begin{IEEEbiography}[{\includegraphics[width=1in,height=1.25in,clip,keepaspectratio]{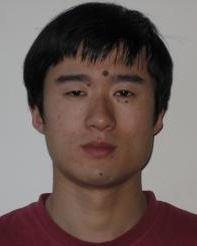}}]{Liangjing Ding}
Liangjing Ding received the BS degree in Electrical Engineering in 2005, the M.S. degree in Biomedical Engineering in 2008, both from University of Science and Technology of China, and the Ph.D degree in Computational Science from Florida State University in 2013. From 2008 to
2013 he was a research assistant under the supervision of professor Adrian Barbu. His research interests are in machine learning and its applications in computer vision.
\vspace{-12mm}
\end{IEEEbiography}

\begin{IEEEbiography}[{\includegraphics[width=1in,height=1.25in,clip,keepaspectratio]{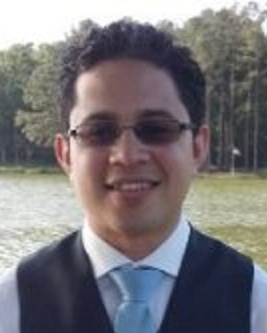}}]{Gary Gramajo} received a BS in Mathematics in 2010, MS in Statistics in 2013, and Ph.D in Statistics in 2015 under the supervision of professor Adrian Barbu, all from Florida State University. In 2014 he was a McKnight Fellow and from 2015 he is a statistician with with Chick-fil-A data analytics in Atlanta, GA. His research interests are in machine learning and its applications to challenging business problems.
\end{IEEEbiography}
\vspace{-12mm}


\clearpage

\section*{Convergence and Consistency Proofs}

{\em Proof.}
We focus on the  classification case. The proof for regression is similar (yet simpler).
To facilitate the derivation, we introduce some necessary notation and symbols. First, we define the \textbf{quantile thresholding} $\Theta^{\#}(\cdot; k, \lambda)$, as a variant of the {hard-ridge} thresholding \cite{SheTISP,She2013Spec}. Given $1\le k \le  M$ and $\lambda\ge 0$, $\Theta^{\#}(\bbeta; k, \lambda): {\mathbb R}^M\rightarrow {\mathbb R}^M$ is defined for any $\bbeta\in {\mathbb R}^M$ such that the $k$ largest components of $\bbeta$ (in absolute value) are shrunk   by a factor of $(1+\lambda)$ and the remaining components are all set to be zero.
In the case of ties, a random tie breaking rule is used.
We write $\Theta^{\#}(\bbeta; k)$ for $\Theta^\#(\bbeta; k, 0)$.
Given a vector $\bsb{\alpha}\in \mathbb R^M$ and a subset $\mathcal S\subset \{1,\cdots, M\}$, $\bsb{\alpha}[\mathcal S]$ or $\bsb{\alpha}_{\mathcal S}$ denotes the subvector of $\bsb{\alpha}$ with the components indexed by $\mathcal S$; similarly, given  a matrix $\bsb{A}\in \mathcal R^{N\times M}$ and  a subset $\mathcal S\subset \{1,\cdots, M\}$,  $\bsb{A}[:, \mathcal S]$ denotes the submatrix of $\bsbX$ formed with the columns indexed by $\mathcal S$.
 Define $$\mu(t) = 1/(1+\exp(-t)). $$ For  a vector $\bsb{t}\in \mathbb R^N$, $\mu(\bsb{t})\in \mathbb R^N$ is defined componentwise.

Recall the design matrix $\bsbX=[\bsbx_1^T, \cdots, \bsbx_N^T]^T\in \mathbb R^{N\times M}$ and the loss $F(\bbeta; \bsbX, \bsb{y})$ (or $F(\bbeta)$ for simplicity) $= \hspace{-1mm}\sum_{i=1}^N (-y_i \bsbx_i^T \bbeta \hspace{-1mm}+ \hspace{-1mm} \log(1 \hspace{-1mm}+ \hspace{-1mm}\exp(\bsbx_i^T \bbeta)))$. Using the previously defined symbols,  it is easy to calculate $\partial F/\partial \bbeta=-\bsbX^T
 \bsb{y} + \bsbX^T \mu(\bsbX \bbeta)$. Likewise,   the Fisher information matrix at $\bbeta$, denoted by
 $\mathcal I(\bsbX, \bbeta)$ with $\mathcal I(\bsbX, \bbeta)[j, k] \triangleq - \mathbb E(\frac{\partial^2 F}{\partial \beta_j \partial \beta_k})$ (where the expectation is taken with respect to $\bsb{y}$), can be explicitly calculated: $ \mathcal I(\bsbX, \bbeta)=\bsbX^T   \mbox{diag}\left\{ \mu(\bsbx_i^T \bbeta) (1-\mu(\bsbx_i^T \bbeta))  \right\}_{i=1}^N \bsbX$.

We   re-state  Steps 3-5 in Algorithm 1  as follows
\begin{align}
&\bbeta^{(e+1)}\leftarrow\hspace{-1mm} \Theta^\#(\bbeta^{(e)}\hspace{-1mm}+ \eta \bsbX^T (
 \bsb{y} - \mu(\bsbX \bbeta^{(e)}));  M_e ) \label{tisp1}\\
&{\mathcal S} \leftarrow \{j: \beta_j^{(e+1)}\neq 0, 1\le j \le M \} \label{tisp2}\\
&\bbeta^{(e+1)}  \leftarrow \bbeta^{(e+1)}[\mathcal S] \label{tisp3}\\
&\bsbX  \leftarrow \bsbX[:, \mathcal S] \label{tisp4}
\end{align}

 \textbf{Part (i):} We show that the following properties hold for any $e$ large enough:
\begin{align}
F(\bbeta^{(e)}) \geq F(\bbeta^{(e+1)}), \mbox{ and } \|\bbeta^{(e)}\|_0 \leq k,
\end{align}
indicating that  FSA  could be viewed as an algorithm for solving
$
\min_\bbeta F(\bbeta)  \mbox{ s.t. } \|\bbeta\|_0\leq k.
$

We begin by analyzing the following sequence of iterates with a fixed threshold parameter  (to make the conclusion more general, we assume varying stepsizes here)
\begin{align}
\bbeta^{(e+1)} &= \Theta^\#(\bbeta^{(e)} + \eta^{(e)} \bsbX^T (\bsb{y} -\mu(\bsbX \bbeta^{(e)})); k ), \label{tisp1b}
\end{align}
where   $\eta^{(e)}$ stands for the step size at the $e$th iteration. Here, $M_e = k \leq M$, but $M$ is possibly larger than $N$. 

\begin{lemma}
Suppose  $0< \eta^{(e)}\leq 4/\| \bsbX \|_2^2$, then for the iterates defined by \eqref{tisp1b}, we have
 $ F(\bbeta^{(e)}) -  F(\bbeta^{(e+1)}) \geq \frac{1/\eta^{(e)} - \|\bsbX\|_2^2/4}{2}\|\bbeta^{(e+1)} - \bbeta^{(e)} \|_2^2$ and $\bbeta^{(e)}$
obeys $\|\bbeta^{(e)}\|_0\leq k$. \label{lem}
\end{lemma}

{\em Proof.}
First define a  surrogate function as follows
\[
\begin{split}
&G(\bbeta, \bsb{\gamma}; \omega) = \sum_i \{- y_i  \bsbx_i^T \bsb{\gamma}   +  \log (1 + e^{\bsbx_i^T \bsb{\gamma}}) \}  + \frac{\omega}{2} \|\bsb{\gamma} - \bbeta\|_2^2
 \\ & - \sum_i \{ \log (1 + e^{\bsbx_i^T \bsb{\gamma}}) - \log (1 + e^{\bsbx_i^T \bbeta})\}
  + \sum_i \frac{\bsbx_i^T \bsb{\gamma}-\bsbx_i^T \bbeta
}{1+e^{-\bsbx_i^T \bbeta}}.
\end{split}
\]
Note that $G$ is quadratic in $\bsb{\gamma}$, and thus given $\bbeta$, minimizing $G$ over $\{\bsb{\gamma}: \|\bsb{\gamma}\|_0\leq k\}$ is equivalent to
$$
\min_\gamma \frac{\omega}{2}  \| \bsb{\gamma} \hspace{-1mm}-\hspace{-1mm} \bbeta \hspace{-1mm}-\hspace{-1mm} \frac{1}{\omega} \bsbX^T \bsb{y} + \frac{1}{\omega} \bsbX^T\hspace{-1mm} \mu(\bsbX \bbeta) \|_2^2 \mbox{ s.t. } \|\bsb{\gamma}\|_{0}\leq k.
$$

It is easy to verify that this problem (though nonconvex)   has a globally optimal solution
obtained by the quantile thresholding (see, e.g.,  \cite{She2013Spec})
$$\bsb{\gamma} = \Theta^\#(\bbeta  + \frac{1}{\omega} \bsbX^T (\bsb{y} -\mu(\bsbX\bbeta)); k).$$

On the other hand, from Taylor expansion, we have \begin{align*}
&   F(\bsb{\gamma}) + \frac{\omega}{2} \|\bsb{\gamma} - \bbeta\|_2^2 - G(\bbeta, \bsb{\gamma}; \omega)  \\&= \sum_i \{ \log (1 + e^{\bsbx_i^T \bsb{\gamma}}) - \log (1 + e^{\bsbx_i^T \bbeta})\}  - \sum_i \frac{\bsbx_i^T \bsb{\gamma}-\bsbx_i^T \bbeta}{1+e^{-\bsbx_i^T \bbeta}} \\
 & = \frac{1}{2} (\bsb{\gamma} - \bbeta)^T \mathcal I(\bsbX, \bsb{\zeta} ) (\bsb{\gamma} - \bbeta),
\end{align*}
where $
\mathcal I(\bsbX, \bsb{\zeta} ) = \bsbX^T   \mbox{diag}\left\{ \frac{e^{\bsbx_i^T \bsb{\zeta}}}{(1+e^{\bsbx_i^T \bsb{\zeta}})^2} \right\}_{i=1}^N \bsbX
$
 with $\bsb{\zeta}=t\bbeta + (1-t) \bsb{\gamma}$ for some $t\in [0, 1]$. Because $\frac{e^{\bsbx_i^T \bsb{\zeta}}}{(1+e^{\bsbx_i^T \bsb{\zeta}})^2} = \frac{1}{1+e^{\bsbx_i^T \bsb{\zeta}}} (1- \frac{1}{1+e^{\bsbx_i^T \bsb{\zeta}}})\le \frac{1}{4}$,  $\frac{1}{2} (\bsb{\gamma} - \bbeta)^T \mathcal I (\bsb{\gamma} - \bbeta) \in \left[0, \frac{1}{2} \frac{\|\bsbX\|_2^2}{4} \|\bsb{\gamma} - \bbeta\|_2^2\right]$.

Finally, for the sequence of  $\bbeta^{(e)}$ defined by \eqref{tisp1b}, we have $\bbeta^{(e+1)} \in \arg\min_{\bsb{\gamma}:\|\bsb{\gamma}\|_{0}\leq k }G(\bbeta^{(e)}, \bsb{\gamma};  1/\eta^{(e)})$ and so
$
G(\bbeta^{(e)}, \bbeta^{(e+1)};  1/\eta^{(e)}) \leq G(\bbeta^{(e)}, \bbeta^{(e)};  1/\eta^{(e)})$. But $G(\bbeta^{(e)}, \bbeta^{(e)};  1/\eta^{(e)})$ is just $ F(\bbeta^{(e)})$, and
$
F(\bbeta^{(e+1)}) + \frac{1}{2 \eta^{(e)}} \|\bbeta^{(e+1)}- \bbeta^{(e)} \|_2^2 - \frac{1}{2} \frac{\|\bsbX\|_2^2}{4} \|\bbeta^{(e+1)} -\bbeta^{(e)} \|_2^2
\leq  G(\bbeta^{(e)}, \bbeta^{(e+1)};  1/\eta^{(e)})$. In summary,   $F(\bbeta^{(e)}) - F(\bbeta^{(e+1)})\geq \frac{1/\eta^{(e)} - \|X\|_2^2/4}{2}\|\bbeta^{(e+1)} - \bbeta^{(e)} \|_2^2$. $\Box$

For the iterates yielded by the original algorithm, notice that  $M_e$ used  in Step \eqref{tisp1} equals $k$  eventually, say, for any $e>E$, and thus $\mathcal S$ and $\bsbX$ stay fixed afterwards. From the lemma, $F(\bbeta^{(e)})$  must be decreasing for $e>E$.

{Remark.} The result implies that in  implementation, we may adopt a universal step choice $\eta$ at any iteration, as long as it satisfies   $\eta \leq 4/ \|\bsbX\|_2^2$.


\textbf{Part (ii):} Now we prove the strict convergence of $\bbeta^{(e)}$ regardless the size  of $M$   (which is a much stronger result than the convergence  of $F(\bbeta^{(e)})$). In fact,    $M\gg N$ may occur.  Without loss of generality, suppose $M_e = k = M$ in \eqref{tisp1b} and thus   \eqref{tisp2}-\eqref{tisp4} are inactive.  Unlike Gaussian regression, logistic regression  may not have a finite maximum likelihood estimator even when restricted to a small set of features, say, when the class labels having $y=0$ and having $y=1$ do not overlap in the predictor space \cite{AlbAnd84}. 
In the literature,   to avoid such irregularities, an  assumption  is usually made based on the following  overlap condition.   (The condition  is however not needed, either in  regression, or when there is an additional $\ell_2$ penalty $\lambda \|\bbeta\|_2^2$ with $\lambda>0$ in regression or logistic regression).

{{Overlap Condition.}} For any  $k$-subset of the $M$ given $x$-features, neither {complete separation} nor {quasicomplete separation} \cite{CAT3}   occurs.

Let $\bar \bbeta$ be a finite solution to
\begin{align}
 \bsbX^T \mu(\bsbX\bbeta) - \bsbX^T \bsb{y} =\bsb{0}, \label{theta-eq}
\end{align}
with the  existence guaranteed under the overlap assumption (cf. \cite{CAT3} for  details). Let the SVD of $\bsbX$ be $\bsbX=\bsb{U} \bsb{D} \bsb{V}^T$ where  $\bsb{D}$ is a $r\times r$ square matrix with all diagonal entries positive and  $r= rank(\bsbX)$. Let $\bsbV_{\perp}$ be an orthogonal complement to $\bsbV$. Construct $\bar\bbeta'=\bsbV\bsbV^T \bar\bbeta + \bsbV_{\perp} \bsbV_{\perp}^T \bbeta^{(0)}$ which still satisfies \eqref{theta-eq}. Without loss of generality,    we  still write $\bar\bbeta'$ as $\bar\bbeta$.

With a bit algebra, $\bbeta^{(e+1)} - \bar\bbeta=\bbeta^{(e)}-\bar\bbeta + \eta (\bsbX^T \bsb{y} -\bsbX^T \mu(\bsbX \bbeta^{(e)})) =\bbeta^{(e)}-\bar\bbeta + \eta (\bsbX^T \mu(\bsbX \bar\bbeta) -\bsbX^T \mu(\bsbX \bbeta^{(e)}))=(\bsb{I}-\eta\mathcal I(\bsbX, \bsb{\zeta}) ) (\bbeta^{(e)}-\bar\bbeta)$ with the last equality again due to    Taylor expansion. However, a  challenge is   that $\bsbX^T  \mbox{diag}\left\{ \mu(\bsbx_i^T \bsb{\zeta}) (1-\mu(\bsbx_i^T \bsb{\zeta}))  \right\}  \bsbX$ may be singular, e.g., when  $M\gg N$ (and thus the iteration mapping is not a contraction). We   introduce $\bsb{\gamma}=\bsb{D}\bsbV^T \bbeta$,  $\bsb{\gamma}^{(e)}=\bsb{D}\bsbV^T \bbeta^{(e)}$, and $ \bar{\bsb{\gamma}}=\bsb{D}\bsbV^T \bar\bbeta$. Then $\bsb{\gamma}^{(e+1)} - \bar{\bsb\gamma} = (\bsb{I} - \eta\bsb{D}^2 \bsb{U}^T  \mbox{diag}\left\{ \mu(\bsbx_i^T \bsb{\zeta}) (1-\mu(\bsbx_i^T \bsb{\zeta}))  \right\} \bsb{U}) (\gamma^{(e)} - \bar{\bsb{\gamma}})$. Under the $\eta$-assumption, $\|(\bsb{I} - \eta\bsb{D}^2 \bsb{U}^T  \mbox{diag}\left\{ \mu(\bsbx_i^T \bsb{\zeta}) (1-\mu(\bsbx_i^T \bsb{\zeta}))  \right\} \bsb{U})\|_2 < c<1$, and so $\bsb{\gamma}^{(e)}$ converges. Nicely, by the construction of $\bar \bbeta$, $\bsbV\bsbV^T (\bbeta^{(e)}-\bar\bbeta)=(\bbeta^{(e)}-\bar\bbeta)$ for any $e$, and thus $\bbeta^{(e)}-\bar\bbeta = \bsbV \bsb{D}^{-1} ({\bsb{\gamma}}^{(e)}-\bar{\bsb{\gamma}})$. This indicates  that the sequence of $\bbeta^{(e)}$   strictly converges although $M$ may be larger than $N$, and the limit point, denoted by $\bbeta_o(N)$,  satisfies \eqref{theta-eq} from which   the local optimality easily follows. 


\textbf{Part (iii):} We construct a two-stage cooling schedule to show the consistency. Recall the assumption  $\mathcal I (\bsbX, \bbeta^*)/ N\rightarrow \mathcal I^*$.    Let $\bbeta_o(N)$ be a solution to \eqref{theta-eq}. From the central limit theorem,  $\sqrt N (\bbeta_o(N) - \bbeta^*) \Rightarrow N(0, {\mathcal I^*}^{-1})$. This indicates that with probability tending to 1,  $\bbeta_o(N)$ is the unique solution.

Stage 1. Set $M_e=M$ ($1\leq e \leq E_{0}$)  for some $E_0$. Accordingly,  the squeezing operations \eqref{tisp2}-\eqref{tisp4} do not take any essential effect.  To bound the estimation error, we write $\|\bbeta^{(e)}- \bbeta^*\|_{\infty} \leq \|\bbeta_o(N) - \bbeta^*\|_{\infty} + \|\bbeta^{(e)} - \bbeta_o(N)\|_{\infty}$. From the central limit theorem, the first term on the right hand side is   $O_p(\frac{1}{\sqrt N})$   and thus $o_p(1)$. Define $\mathcal S^*=\{j: \beta_j^*\neq 0\}$ and $\min |\bbeta_{\mathcal S^*}^*|=\min_{j\in \mathcal S^*} |\beta_j^*|$. Then  $\|\bbeta_o(N) - \bbeta^*\|_{\infty}\leq \frac{1}{8} \min |\bbeta_{\mathcal S^*}^*|$ with probability tending to 1 as $N\rightarrow \infty$.

On the other hand, from the algorithmic convergence of $\bbeta^{(e)}$ established above, there exists $E_{1}$ (or $E_{1}(N)$, as a matter of fact) large enough such that $\|\bbeta^{(e)} - \bbeta_o(N)\|_{\infty}\leq \frac{1}{8} \min |\bbeta_{\mathcal S^*}^*|$ and $\| \bsbX^T \mu(\bsbX\bbeta^{(e)}) - \bsbX^T \bsb{y}\|_{\infty}\leq \frac{1}{8 \eta} \min |\bbeta_{S^*}^*|$, $\forall e \geq E_{1}$. Therefore, if we choose $E_0\geq E_1$,  $\|\bbeta^{(e)}-\bbeta^*\|_{\infty}\leq \frac{1}{4} \min |\bbeta_{\mathcal S^*}^*|$ and $\| \bsbX^T \mu(\bsbX\bbeta^{(e)}) - \bsbX^T \bsb{y}\|_{\infty}\leq \frac{1}{8  \eta} \min |\bbeta_{\mathcal S^*}^*|$, for $\forall e: E_1\leq  e\leq E_0$ occur with probability tending to 1.

Stage 2. Now set $M_e=k\geq  \|\bbeta^*\|_0$, $\forall e > E_0$. At $e=E_0+1$, letting $\bsb{\xi}=\bbeta^{(e)}  +\eta \bsbX^T (\bsb{y} -\mu(\bsbX\bbeta^{(e)}))$, we have $\min |\bsb{\xi}_{\mathcal S^*}| \geq \frac{5}{8} \min |\bbeta_{\mathcal S^*}^*| > \frac{3}{8} \min |\bbeta_{\mathcal S^*}^*| \geq\max |\bsb{\xi}_{\mathcal {S^*}^c}|$. Therefore, Step \eqref{tisp1}  yields $\beta_{j}^{(e+1)}\neq 0$, $\forall j \in \mathcal S^*$. After Steps \eqref{tisp2}-\eqref{tisp4}, all relevant predictors are kept  with probability tending to 1. Repeating the argument in Stage 1 gives the consistency of $\bbeta^{(e)}$ for any $e$ sufficiently large.

 The proof in the Gaussian case follows the same lines. But in Part (ii) we do not have to assume the overlap condition, 
 due to Landweber's classical convergence result \cite{Land51}.

\vspace{-1mm}
\section*{ Ranking for Motion Segmentation}

A major difficulty with spectral clustering is that a rigid motion lies in a low dimensional space that does not have a fixed dimension. As a result, when there are several motions present in the same video sequence, it is hard to determine the best projection
dimension for spectral clustering.
Consequently, some segmentation methods~\cite{ding2012motion,lauer2009spectral} propose
to project to a number of spaces of different dimensions and find the best results according to some measure.

However, it is hard to find a versatile measure that consistently finds the best dimension in all scenarios.
Moreover, segmentation algorithms always have one or more parameters, such as the
noise level, the separability of the affinity measure, etc,
that need to be tuned according to different scenarios.
It is also hard to expect there exists a set of parameters that work well for all problems.

Furthermore, many motion segmentation algorithms have been published in recent years, each with their own strength and weaknesses.
It would be of  practical importance to segment one sequence by many
different algorithms and find an automatic way to select the best segmentation.

In this work, we address the problem of choosing the best segmentation from a larger set of segmentations that are generated
by different algorithms or one algorithm with different parameters. We formalize it as a ranking problem and solve it using supervised learning with the FSA-Rank algorithm.

\subsection*{Segmentation by Spectral Clustering}
The candidate segmentation results are generated by the velocity clustering (VC)
algorithm~\cite{ding2012motion}, which we briefly describe below to make the paper self-contained.

A trajectory $t=[(x_1,y_1),...,(x_F,y_F)]$ is transformed into a velocity vector
\begin{equation}
v(t)=[x^1 - x^2\hspace{-1mm}, y^1 - y^2\hspace{-1mm}, \ldots,  x^{F-1} - x^{F}\hspace{-1mm}, y^{F-1} - y^{F}\hspace{-1mm}, x^F\hspace{-1mm}, y^F]^T
\label{eq:velocity}
\end{equation}
where $F$ is the number of frames of the image sequence.  Then the velocity vectors are projected to spaces of
different dimensions in range $[2K, 4K]$ by truncated SVD, where $K$ is the
number of motions.
The range contains the possible dimensions of spaces containing $K$ mixed rigid motions. At last, spectral
clustering is applied to obtain the segmentation using the angular affinity
\begin{equation}
  A_{ij} = (\frac{t_i't_j}{\|t_i\| \|t_j\|})^{2\alpha},\;  i,j\in \{1,...,P\}
  \label{angular-measure}
\end{equation}
where $t_i$ and $t_j$ are two projected trajectories, and $\alpha$ is a tuning
parameter to improve inter-cluster separability. In this paper the value $\alpha$ is set to 2, as in
VC \cite{ding2012motion}. Please refer to \cite{ding2012motion} for more details.

After removing possible repetitive segmentations, around $2K+1$ segmentations
would be generated for each sequence.
While the VC method proposes an error measure to select the best
segmentation, this paper solves the same problem by learning.

\subsection*{Likelihood and Prior Based Features}

A motion segmentation can be described by a labeling $L:\{1,..,P\}\to \{1,..,K\}$. We will use two types of features that can characterize the ranking of a motion segmentation $L$: likelihood features and prior features.

Under the orthographic camera assumptions, the point trajectories of each rigid motion should lie in a 3 dimensional affine subspace.

For a segmentation the {\bf likelihood features} are used to measure how far are the point trajectories of the same label from lying in a 3D linear subspace.

For both the original trajectory vectors and the points obtained by projection to space of dimension $d$, where $d$ is a parameter, we fit in a least squares sense 3-D
affine subspaces $S_l$ through the points of motion label $l\in \{1,...,K\}$. Denote $L(i)$ as the label of trajectory $t_i$  and let $D(t, S)$ be the euclidean distance of point $t$ to plane $S$. Let $N$ is the total number of trajectories.

We use three types of likelihood features:
\begin{itemize}
\item The average distance$
\frac{1}{N}\sum_{i=1}^ND(t_i, S_{L(i)})
$
\item The average squared distance$
\frac{1}{N}\sum_{i=1}^ND^2(t_i, S_{L(i)})
$
\item The average thresholded distance
\vspace{-2mm}
\[
\frac{1}{N}\sum_{i=1}^NI(D(t_i, S_{L(i)})\ge \tau),
\vspace{-2mm}
\]
where $I(\cdot)$ is the indicator function taking on value 1 if its argument is true
or 0 otherwise, and $\tau$ is a threshold.
\end{itemize}

Inspired by VC, the first and second types of features obtained in all dimensions $d\in [2K, 4K]$ are sorted and the smallest 4 values are used as
features.

By changing the threshold $\tau$ and  dimension $d$ a number of features of the third type can be obtained.

The {\bf prior features} measure the compactness of the partition over different graphs.

For a given $k$, the $k$-nearest neighbor (kNN) graph is constructed using a distance
measure in a space of a given dimension $d$. The distance could be either the Euclidean
distance or the angular distance defined in eq.~\ref{angular-measure}. By changing the dimension $d$, number of neighbors $k$ and distance measure a number of different graphs and features are obtained.

On the kNN graph $G=(V,E)$ the prior feature is the proportion of the edges that connect vertices with different labels
\vspace{-2mm}
\[
F_G=\frac{|(i,j)\in E, L(i)\not =L(j)|}{|E|}
\vspace{-2mm}
\]
where $L(i)$ is the segmentation label of vertex $i\in V$.

In total, the features described in this section result in more than 2000 features for each segmentation.

\vspace{-3mm}
\subsection*{Motion Segmentation Algorithm}

Given a new sequence, the learned parameter vector $\bbeta$ is used to select the best segmentation for that sequence. The whole procedure is described in Algorithm \ref{alg:vsc}.
\vskip -2mm
\begin{algorithm}[ht]
  \caption{{\bf Motion Segmentation using Ranking}}
  \label{alg:vsc}
\vskip -0mm
\begin{algorithmic}
  \STATE {\bf Input: } The measurement matrix $W = [t_1, t_2, \ldots, t_P] \in \mathbb{R}^{2F\times P}$ whose columns
  are point trajectories, and the number of clusters $K$.

\STATE {\bf Preprocessing: } Build the velocity measurement matrix
$W'=(v(t_1),...,v(t_P))$ where $v(t)$ is given in eq. \eqref{eq:velocity}.

\FOR{$d= d_{min}$ {\bf  to} $d_{max}$}

\STATE 1. Perform SVD:  $W' = U\Sigma V^T$
\STATE 2. Obtain $P$ projected points as the columns of the $d\times P$ matrix
\vspace{-2mm}
\[X_d  = [v_1, ... , v_d]^T
\vspace{-2mm}
\] where $v_i$ is the $i$-th column of $V$.

\STATE 3. Apply spectral clustering to the $P$ points of $X_d$ using the affinity measure \eqref{angular-measure}, obtaining segmentation $L_d$.
\STATE 4. Extract feature vector $\bx_d$ from segmentation $L_d$ as described above.
\STATE 5. Compute the ranking
\vspace{-2mm}
\[
f_\bbeta(L_d)= \sum_{k=1}^M  \bu_k^T(x_{dk})\bbeta_k
\vspace{-2mm}
\]

  \ENDFOR

\STATE  {\bf Output:} The segmentation result $L_d$ with the largest value of $f_\bbeta(L_d)$.
\end{algorithmic}
\end{algorithm}

\subsection*{Training the Ranking Function}

The performance of a segmentation is
characterized by the misclassification error
\vspace{-2mm}
\begin{equation}
  \textrm{Misclassification Error} = \frac{\textrm{\# misclassified points}}{\textrm{total \# of points}},
\vspace{-2mm}
\end{equation}
which could be easily calculated by comparison to the ground truth
segmentation.

The true rankings $r_{ij}, (i,j)\in C$ are constructed based on the relative misclassification errors of the
segmentations. Since at test time only the segmentations belonging to the same sequence will be compared, the set $C$ contains only pairs of segmentations obtained from the same sequence.

For any two segmentations $i,j$ obtained from the same sequence, the ranking $r_{ij}$ is based on the misclassification errors of the two segmentations, with value $1$ is $i$ is better than $j$, $0.5$ if they have the same error and $0$ if $j$ is better than $i$.

These ground truth rankings and the feature vectors for each segmentation are used in the FSA-Rank Algorithm to obtain the parameter vector $\bbeta$.

\subsection*{The RankBoost Algorithm}
The RankBoost algorithm~\cite{freund2003efficient} is used in this paper as
a baseline method to compare performance in learning the ranking function.

Let $S=\{\bx_i\in \RR^M,
i=\overline{1,N}\}$ be the set of training instances. We assume that a ground
truth ranking is given on a subset $C\subset \{1,...,N\}\times \{1,...,N\}$ as $r_{ij},(i,j)\in C$
where $r_{ij}> 0$ means $x_i$ should be ranked above $x_j$ and vice
versa.

RankBoost searches for a ranking which is similar to the given ranking $r$. To formalize the goal, a distribution $D$ is constructed by $D_{ij} = c \cdot \max\{0, r_{ij}\}$, where $c$ is a constant to make $\sum_{(i,j)\in C}D_{ij} = 1$. The learning algorithm tries to find a ranking function $H: \RR^M \rightarrow \RR$ that minimizes the weighted sum of  wrong orderings:
\vspace{-2mm}
\begin{equation*}
\textrm{rloss}_D = \sum_{(i,j)\in C} D_{ij}I(H(\bx_1) \le H(\bx_0))
\vspace{-2mm}
\end{equation*}
where again $I(\pi)$ is 1 if predicate $\pi$ holds and 0
otherwise. The ranking function $H(x)$ is a weighted sum of weak rankers that
are selected iteratively
\vspace{-2mm}
\begin{equation*}
  H(\bx) = \sum_{t = 1}^T\alpha_th_t(\bx).
\vspace{-2mm}
\end{equation*}
At iteration $t$, RankBoost selects the best weak ranker $h_t$ along with its
weighted ranking score $\alpha_t$ from the pool of candidate weak rankers, and adds
$\alpha_th_t(x)$ to the ranking function $f_{t-1}(x)$.

We used threshold-based weak rankers
\vspace{-1mm}
\begin{equation}
  h(\bx) = \left\{ \begin{array}{ll}
        1 & \textrm{\quad if } x_i > \theta \\
        0 & \textrm{\quad if } x_i \le \theta
        \end{array} \right.
\label{eqn:weak-ranker}
\vspace{-1mm}
\end{equation}
that depend on the threshold $\theta \in \mathbb{R}$ and the variable index $i$. The pool of weak rankers is generated using all variables $i=\overline{1,M}$ and $B=64$ equally spaced thresholds on the range of each feature.

\end{document}